\documentclass[twoside]{article}

\usepackage[accepted]{aistats2021}
\usepackage[round]{natbib}

\usepackage{graphicx}
\usepackage{amsfonts} 
\usepackage{amsmath}
\usepackage{amssymb}
\usepackage{mathtools}
\usepackage{subfigure}
\usepackage{hyperref}

\newcommand\mysim{\stackrel{\mathclap{\normalfont\mbox{\tiny iid}}}{\sim}}
\DeclareMathOperator*{\elbo}{ELBO}

\DeclareMathOperator*{\kl}{KL}

\begin{document}

%
\runningtitle{Latent Gaussian process with composite likelihoods and numerical quadrature}

%
\runningauthor{Ramchandran, Koskinen, L\"ahdesm\"aki}

\twocolumn[

\aistatstitle{Latent Gaussian process with composite likelihoods\\ and numerical quadrature}


\aistatsauthor{ Siddharth Ramchandran\textsuperscript{1} \quad Miika Koskinen\textsuperscript{2,3} \quad Harri L\"ahdesm\"aki\textsuperscript{1}}

\aistatsaddress{\\\textsuperscript{1}Department of Computer Science, Aalto University, Finland\\\textsuperscript{2}HUS Helsinki University Hospital, Finland\\ \textsuperscript{3}Faculty of Medicine, University of Helsinki, Finland\\ \href{mailto:siddharth.ramchandran@aalto.fi}{\nolinkurl{siddharth.ramchandran@aalto.fi}}} ]

\begin{abstract}
Clinical patient records are an example of high-dimensional data that is typically collected from disparate sources and comprises of multiple likelihoods with noisy as well as missing values. In this work, we propose an unsupervised generative model that can learn a low-dimensional representation among the observations in a latent space, while making use of all available data in a heterogeneous data setting with missing values. We improve upon the existing Gaussian process latent variable model (GPLVM) by incorporating multiple likelihoods and deep neural network parameterised back-constraints to create a non-linear dimensionality reduction technique for heterogeneous data. In addition, we develop a variational inference method for our model that uses numerical quadrature.
We establish the effectiveness of our model and compare against existing GPLVM methods on a standard benchmark dataset as well as on clinical data of Parkinson's disease patients treated at the HUS Helsinki University Hospital. 
\end{abstract}

\section{Introduction}
Gaussian process (GP) models are flexible probabilistic models that can  perform tasks such as classification and regression, and are popular algorithms in machine learning \citep{rasmussen2004gaussian}. \cite{lawrence2004gaussian} reinterpreted principal component analysis (PCA) as a GP mapping from the latent space to the data space and proposed a generalisation by using a prior that allows for non-linear embedding. This is called the Gaussian process latent variable model (GPLVM). In short, the GPLVM attempts to learn a smooth mapping from the latent space to the data space.

To accurately capture the latent manifold structure of the data, it is important for a dimensionality reduction algorithm to balance between preserving the distance between nearby data points and ensuring that data points that are distant in the data space are not nearby in the latent space (dissimilarity). However, the GPLVM algorithm only guarantees the latter and does not have any constraint that ensures the former. \cite{lawrence2006local} discusses this issue in detail and introduces the idea of incorporating a local-distance preserving constraint thereby formulating a back-constrained GPLVM. \cite{bui2015stochastic} imposed this constraint by using recognition models (or neural networks).

GPLVMs are targeted towards homogeneous datasets (i.e.\ data from a single observation space or likelihood). This poses a significant challenge in our setting where different data items can have different likelihoods. \cite{shon2006learning} proposed a generalisation of the GPLVM model that can handle multiple observation spaces (albeit with Gaussian likelihoods) where the observation spaces are linked by a lower dimensional latent variable space. This was extended by \cite{ek2007gaussian} for three-dimensional human pose estimation by incorporating constraints to the latent space. We build upon the idea of obtaining a shared latent space or a common low-dimensional latent representation using a shared GPLVM as proposed in \cite{ek2007gaussian}. In particular, we extend the idea of shared GPLVM to support multiple likelihoods. Here, the use of non-Gaussian likelihoods introduces intractability into the inference. \cite{titsias2010bayesian} introduced variational inference to the GPLVM  assuming the standard Gaussian noise model. However, this model cannot be extended to multiple likelihoods due to the lack of an analytical solution for the optimal variational distribution. We overcome this by using a sampling-based variational inference with numerical integration by Gauss-Hermite quadrature.

Variational inference seeks to approximate the true posterior distribution by minimising the Kullback-Leibler divergence between the true posterior and a surrogate distribution. 
\cite{hoffman2013stochastic} improved the efficiency of variational inference by proposing an algorithm called stochastic variational inference that incorporated stochastic optimisation into variational inference. To overcome the intractability in our setting, we make use of a variant of stochastic variational inference, called sampling-based variational inference \citep{rezende2014stochastic, kingma2013auto, titsias2014doubly}, and combine that with numerical quadrature. The faster convergence achieved by using mini-batching with the recognition models compensates for the sampling overheads. The introduction of the recognition models brings our method closer to the variational autoencoder. Autoencoders try to learn a latent representation using a neural network to encode the data from the data space to a low-dimensional latent space (encoder) and a separate neural network to decode the data from the low-dimensional latent space back to the data space (decoder) \citep{hinton2006reducing}. In our approach, the recognition model introduced into the extended GPLVM architecture acts as a form of encoder, while the probabilistic GP mapping acts as a decoder.

\begin{figure}[!t]
\centering
\includegraphics[width=0.8\linewidth]{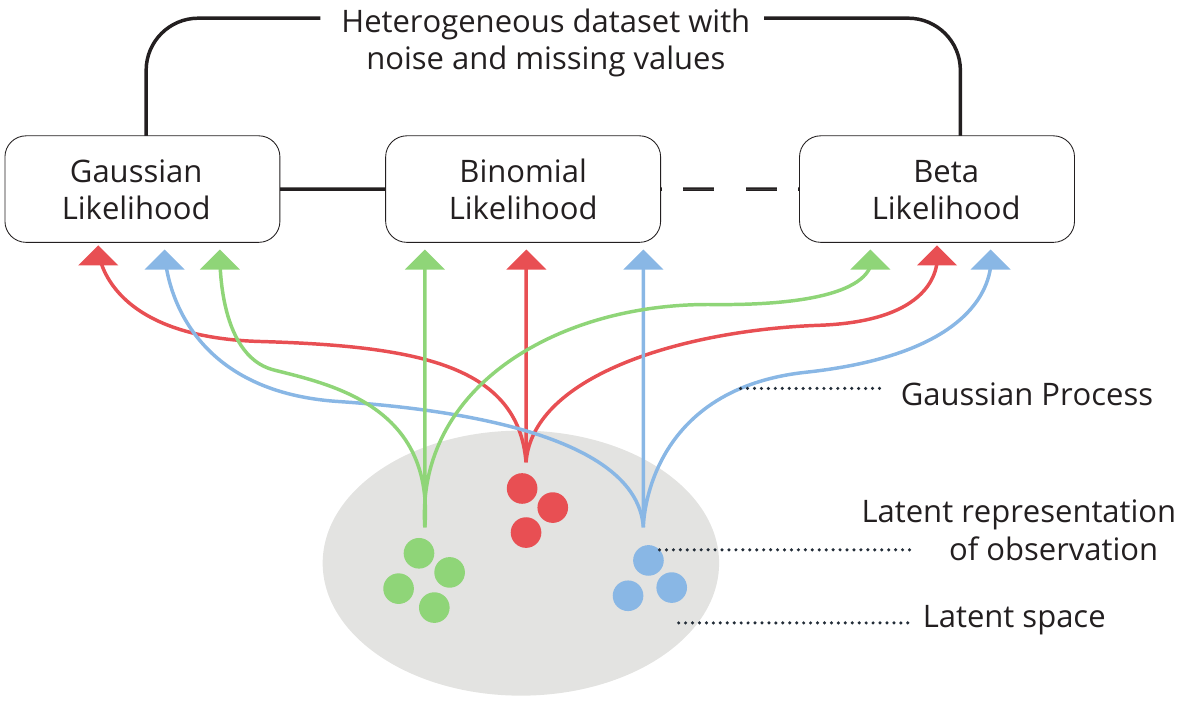}
\caption{Model overview. Each point in the image corresponds to an observation comprising of measurements from different likelihoods, that has been projected on to a two dimensional latent space. The colour coding in the latent space would correspond to cluster membership.}
\label{fig:simp_overview}
\end{figure}

Our main contributions in this paper are:
\begin{itemize}
    \item An extension of the GPLVM to produce low-dimensional embeddings of heterogeneous datasets with missing values. We achieve that by linking the stochastic outputs of latent Gaussian processes to modulate the parameters of the different likelihoods through the use of link functions.
    \item We derive a variational lower bound ($\elbo$) that makes use of numerical quadrature and is suitable for stochastic optimisation as in \citep{gal2015latent}.
    \item We make use of the idea of back-constraints to parameterise the variational inference in order to encourage distant points in the data space to be distant in the latent space while preserving local similarities \citep{lawrence2006local}, and allow minibatching that scales GPLVMs to large-scale datasets as in  \citep{bui2015stochastic}.
\end{itemize}
We demonstrate the applicability of our proposed method on clinical data of patients treated for Parkinson's disease at the HUS Helsinki University Hospital as well as on a simulated heterogeneous dataset. Fig.~\ref{fig:simp_overview} illustrates our study's objective.

\section{Methods}
\label{Methods}
Consider a generative model for a dataset $\boldsymbol{Y} = (\boldsymbol{y}_1,\ldots,\boldsymbol{y}_N)^T$ with $N$ observations (or patients in our case) and $D$ variables of possibly different observation spaces (or as in our case, patient records from several disparate sources). The dataset can be represented by a set of output functions $\mathcal{Y} = \{y_d(\boldsymbol{x}_n)\}_{d=1}^D$, where $\boldsymbol{x}_n \in \mathbb{R}^Q$ is the $Q$ dimensional latent space representation for the $n^{\text{th}}$ observation $\forall n = 1,...,N$. Every observation (row) in $\boldsymbol{Y}$ can be represented by a $Q$ dimensional $ \boldsymbol{x}$, and collectively $\boldsymbol{Y}$ can be represented by $\boldsymbol{X} = (\boldsymbol{x}_1,\ldots,\boldsymbol{x}_N)^T \in \mathbb{R}^{N \times Q}$. The traditional GPLVM model considers the case where $y_d(\boldsymbol{x})$ is Gaussian distributed \citep{lawrence2004gaussian}. \cite{wu2017gaussian} and \cite{gal2015latent} have proposed modifications to the GPLVM for Poisson and categorical data, respectively. Similarly, in the supervised learning setting, \cite{moreno2018} proposed an extension of the multi-output Gaussian process regression that can handle heterogeneous outputs. Our work is also close to the mixed likelihood GPLVM model proposed by \citet{murray2018mixed}. However, our method involves heterogeneous likelihood models together with back-constraints and numerical integration that allows us to develop a model inference that can utilise stochastic gradient based optimisation with mini-batching, which has a significant impact on computation efficiency and stability as well as the robustness and quality of learnt latent space. We also demonstrate the utility of such a model on a real-world clinical dataset.

In this paper, we propose a shared GPLVM \citep{ek2007gaussian} approach for which data items in  $\boldsymbol{Y}$ may be differently distributed following Gaussian, binary, beta, Poisson or categorical distributions. We assume that the likelihood for the $d$\textsuperscript{th} variable, $y_d(\boldsymbol{x}_n)$, is specified by a set of parameters $\boldsymbol{\vartheta}_d(\boldsymbol{x}_n) = [\vartheta_{d,1}(\boldsymbol{x}_n),\ldots,\vartheta_{d,P_d}(\boldsymbol{x}_n)] \in \mathcal{\psi}^{P_d}$, where $P_d$ is the number of parameters that define the distribution and $\mathcal{\psi}$ is a generic domain for the parameters. We can think of each element $\vartheta_{d,p}(\boldsymbol{x_n})$ of parameter vector $\boldsymbol{\vartheta}_d(\boldsymbol{x}_n)$ as a non-linear transformation of a Gaussian process prior $\mathcal{F}_{d,p}$, such that $\vartheta_{d,p}(\boldsymbol{x}_n) = \phi_{d,p}(\mathcal{F}_{d,p}(\boldsymbol{x}_n))$ where $\phi_{d,p}(\cdot)$ acts as a link function (deterministic function) that maps the GP output to the appropriate domain for the parameter $\vartheta_{d,p}$.

\begin{figure}[!t]
\centering
\includegraphics[width=0.6\linewidth]{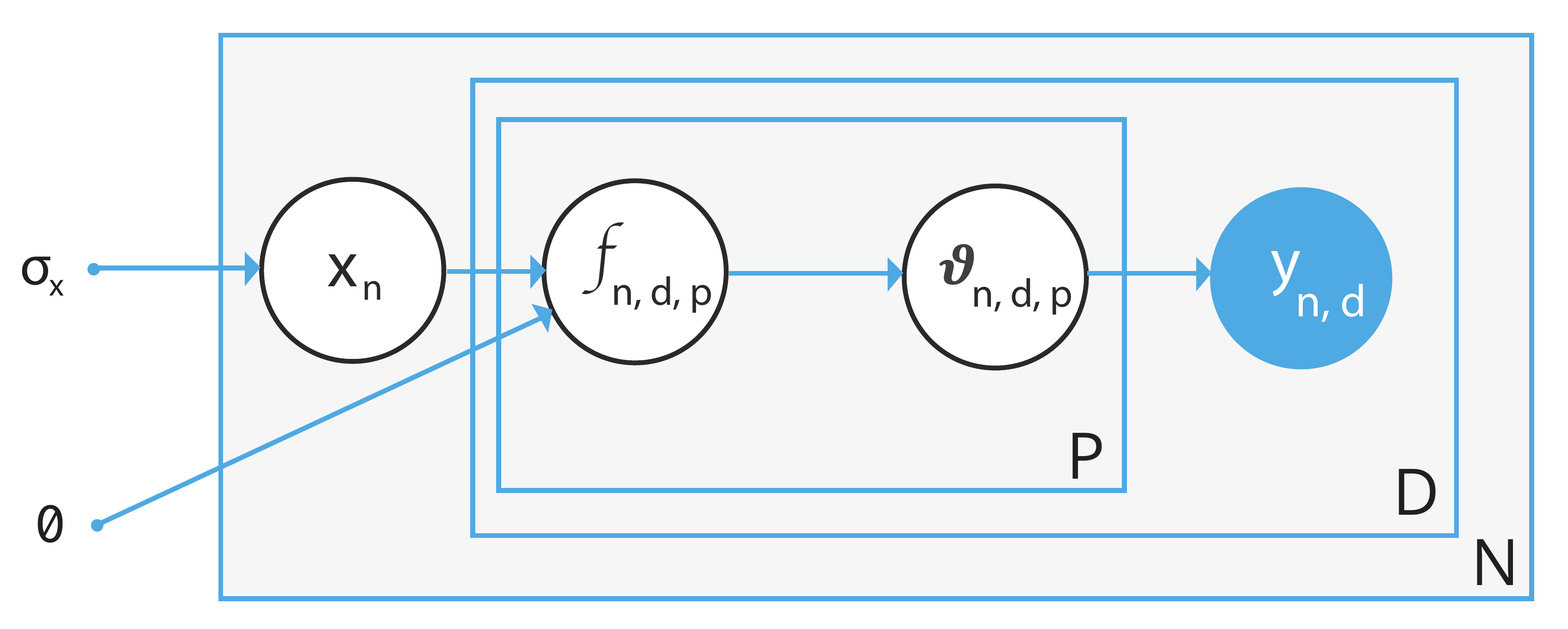}
\caption{Plate diagram of the model. $N$, $D$ and $P$ corresponds to the number of observations, variables, and parameters respectively. The shaded circle refers to an observed variable and the un-shaded circles corresponds to un-observed variables. The learnt hyper-parameters are depicted by $\theta$.}
\label{fig:plate}
\end{figure}
To complete the generative model, we assign a Gaussian distribution prior with standard deviation $\sigma^2_x$ for the latent variables $\boldsymbol{x}_n = (x_{n,1},\ldots,x_{n,Q})^T$. The model can be described by the following equations:
\begin{align}
    x_{n,q} &\mysim \mathcal{N}(0, \sigma^2_x) \\
    \mathcal{F}_{d,p} &\mysim \mathcal{GP}(0, k_d(\cdot)) \\
    f_{n,d,p} &= \mathcal{F}_{d,p}(\boldsymbol{x}_n) \label{eq:GPrealizations}\\
    \vartheta_{d,p}(\boldsymbol{x}_n) &= \phi_{d,p}(f_{n,d,p})\\
    y_{n,d} &\sim p(\cdot| \boldsymbol{\vartheta}_d(\boldsymbol{x}_n)),
\end{align}
where $n \in \{1,\ldots,N\}$, $q \in \{1,\ldots,Q\}$, $d \in \{1,\ldots,D\}$, $p \in \{1,\ldots,P_d\}$, 
$k_d(\cdot)$ is the GP kernel function, and $p( \cdot | \boldsymbol{\vartheta}_d(\boldsymbol{x}))$ denotes a generic likelihood function for the $d$\textsuperscript{th} variable. Our model uses the automatic relevance determination radial basis function as the kernel $k_d(\cdot)$. 

To make the notation concrete, let us consider a case where each observation is comprised of two likelihoods and $D=4$. Let the first two variables be Gaussian distributed and the last two correspond to count data which we assume to follow a Poisson distribution. In other words, $\mathcal{Y} = \{y_1(\boldsymbol{x}), y_2(\boldsymbol{x}), y_3(\boldsymbol{x}), y_4(\boldsymbol{x})\}$ where $y_1(\boldsymbol{x})$ and $y_2(\boldsymbol{x})$ are Gaussian distributed and $y_3(\boldsymbol{x})$ and $y_4(\boldsymbol{x})$ are Poisson distributed. We can say that $y_1(\boldsymbol{x})$ is modelled by two sets of parameters ($P_1 = 2$), $\boldsymbol{\vartheta}_1(\boldsymbol{x}) = [\vartheta_{1,1}(\boldsymbol{x}) \quad \vartheta_{1,2}(\boldsymbol{x})]$ corresponding to the mean and variance, which are functions of $\boldsymbol{x}$ respectively. We can re-write this as $\boldsymbol{\vartheta}_1(\boldsymbol{x}) = [\phi_{1,1}(\mathcal{F}_{1,1}(\boldsymbol{x})) \quad  \phi_{1,2}(\mathcal{F}_{1,2}(\boldsymbol{x}))]$ where $\phi_{1,1}(\cdot)$ would be the identity function and $\phi_{1,2}(\cdot)$ could be the exponential function to ensure that variance takes strictly positive values. Likewise, $y_2(\boldsymbol{x})$ would have a similar formulation. On the other hand, $y_3(\boldsymbol{x})$ and $y_4(\boldsymbol{x})$ would be modelled by the Poisson distribution which uses a single parameter ($P_3 = 1$) corresponding to the event rate (also written as $\lambda$). The outputs of $y_3(\boldsymbol{x})$ and $y_4(\boldsymbol{x})$ correspond to count variables that can take values, $y_3(\boldsymbol{x}),\: y_4(\boldsymbol{x}) \in \mathbb{N} \cup \{0\}$. Considering just $y_3(\boldsymbol{x})$ for now, we can say that it is modelled by $\boldsymbol{\vartheta}_3 = \vartheta_{3,1}(\boldsymbol{x}) = \phi_{3,1}(\mathcal{F}_{3,1}(\boldsymbol{x}))$. The rate parameter is restricted to positive real numbers, hence $\phi_{3,1}(\cdot)$ could be the exponential function that maps $\exp : \mathbb{R} \rightarrow (0, \infty)$. Likewise, $y_4(\boldsymbol{x})$ would have a similar formulation.

For our model, we assume that the outputs are conditionally independent given the vector of parameters denoted by $\boldsymbol{\vartheta}(\boldsymbol{x}) = [\boldsymbol{\vartheta}_1(\boldsymbol{x}), \boldsymbol{\vartheta}_2(\boldsymbol{x}), \boldsymbol{\vartheta}_3(\boldsymbol{x}), \ldots , \boldsymbol{\vartheta}_D(\boldsymbol{x})]$. Hence, the composite likelihood can be defined as
\begin{equation*}
p(\boldsymbol{y}(\boldsymbol{x}) | \boldsymbol{\vartheta}(\boldsymbol{x})) = p(\boldsymbol{y}(\boldsymbol{x}) | \boldsymbol{f}(\boldsymbol{x})) = \prod^D_{d=1}p(y_d(\boldsymbol{x}) | \boldsymbol{\vartheta}_d(\boldsymbol{x})),
\end{equation*}
where $\boldsymbol{f}$ contains realisations of all the GPs from Eq.~\eqref{eq:GPrealizations}. Previous works assume that all the variables are from the same observation space. In other words, a homogeneous dataset was represented by a single likelihood. We generalise the GPLVM model to $D \geq 1$ with possibly different likelihoods, thereby allowing it to create low-dimensional representations of heterogeneous datasets (or data from different observation spaces) that are represented by several different likelihoods while capturing the similarities between the observations. Fig.~\ref{fig:plate} illustrates our model as a plate diagram.
\subsection{Likelihood models}
We consider the cases of Gaussian, binomial, beta, Poisson, and categorical distributions in our analysis. Our model can be easily extended to other distributions as well.
\paragraph{Gaussian distribution}
For the Gaussian distribution, the distribution is specified by two parameters: mean and variance. The mean for each data point is obtained from the GPs, while the variance is a shared parameter that is optimised (and constrained to a positive value) to minimise the computational overhead. For the $d^{\text{th}}$ measured variable this can be written as $\boldsymbol{\vartheta}_d = [\vartheta_{d,1}(\boldsymbol{x})]$ where $\vartheta_{d,1}(\boldsymbol{x})$ is the mean. Therefore, the mean is given by $\vartheta_{d,1}(\boldsymbol{x}) = \phi_{d,1}(\mathcal{F}_{d,1}(\boldsymbol{x}))$ where we choose $\phi_{d,1}(\cdot)$ to be the identity function.
\paragraph{Binomial distribution}
A binomial distributions is specified by two parameters: number of trials and probability of success in each trial. In our case, for each data point the number of trials is 1. Hence, this can be considered as a Bernoulli trial. We can write  $\vartheta_{d,1}(\boldsymbol{x})$ as the probability of success for the $d^{\text{th}}$ variable such that $\boldsymbol{\vartheta}_d = [\vartheta_{d,1}(\boldsymbol{x})]$. The probability of success would be given by $\vartheta_{d,1}(\boldsymbol{x}) = \phi_{d,1}(\mathcal{F}_{d,1}(\boldsymbol{x}))$ where we choose $\phi_{d,1}(\cdot)$ to be the \emph{sigmoid} function (or \emph{softmax} if considering success and failure separately).
\paragraph{Beta distribution}
We re-parameterise the beta distribution in terms of mean, $\mu$. Therefore, the two positive shape parameters ($\alpha$ and $\beta$) can be written as $\alpha = \nu \mu$ and $\beta = \nu (1-\mu)$, 
where $\nu$ is the inverse dispersion parameter which is a shared parameter that is optimised (and constrained to a positive value). Similar to the previous distributions, $\mu$ for each data point is given by $\vartheta_{d,1}(\boldsymbol{x}) = \phi_{d,1}(\mathcal{F}_{d,1}(\boldsymbol{x}))$ where we choose $\phi_{d,1}(\cdot)$ to be the CDF of the standard normal distribution (i.e.\ $\phi_{d,1}(\mathcal{F}_{d,1}(\boldsymbol{x})) = \Phi(\mathcal{F}_{d,1}(\boldsymbol{x}))$).

\paragraph{Poisson distribution}
The Poisson distribution is specified by a single positive parameter known as the rate parameter ($\lambda$). Similar to the previous distributions, $\lambda$ for each data point is given by $\vartheta_{d,1}(\boldsymbol{x}) = \phi_{d,1}(\mathcal{F}_{d,1}(\boldsymbol{x}))$ where we choose $\phi_{d,1}(\cdot)$ to be the exponential function. 
\paragraph{Categorical distribution}
For the categorical distribution, we make use of a formulation similar to \cite{gal2015latent} which is a generalisation of the binomial distribution (see Sec.\ 2 in Suppl.\ Material).

\subsection{Auxiliary variables}
The computational complexity of the Gaussian process models is reduced by the introduction of auxiliary variables or inducing inputs \citep{titsias2009variational}. We consider a set of $M$ inducing inputs, $\boldsymbol{Z} = (\boldsymbol{z}_1,\ldots,\boldsymbol{z}_M)^T \in \mathbb{R}^{M \times Q}$ that lie in the $Q$ dimensional latent space. Their corresponding outputs in the input space would be $\boldsymbol{U} = (\boldsymbol{u}_1,\ldots,\boldsymbol{u}_M) \in \mathbb{R}^{M \times D}$. According to \cite{quinonero2005unifying}, the auxiliary variables act as a support for the covariance function of the GP thereby allowing it to be evaluated on these points instead of the entire dataset.  Hence, we can perform approximate inference in a time complexity of $\mathcal{O}(M^2N)$ instead of $\mathcal{O}(N^3)$ by evaluating the covariance function of the GP on the auxiliary variables instead of the entire dataset. Continuing the model description, we can write $u_{m,d} = \mathcal{F}_{d,p}(\boldsymbol{z}_m)$. Moreover, the joint distribution of $(\boldsymbol{f}_d, \boldsymbol{u}_d)$ is a multi-variate Gaussian distribution $N(0, \boldsymbol{K}_d([\boldsymbol{X}, \boldsymbol{Z}],[\boldsymbol{X}, \boldsymbol{Z}]))$. Further marginalising the inducing outputs leads to a joint distribution of the form $\boldsymbol{f}_d \sim \mathcal{N}(0, \boldsymbol{K}_d(\boldsymbol{X}, \boldsymbol{X})), \forall d$ such that $\boldsymbol{f}_d \in \mathbb{R}^{N \cdot P_d \times 1}$ (note that, except for the categorical distribution, $P_d=1$). Hence, the marginal likelihood of the data remains unchanged by the introduction of the auxiliary variables.

\subsection{Variational inference}
In our model, the marginal log-likelihood is intractable due to the presence of an arbitrary number of non-Gaussian likelihoods. Hence, we make use of variational inference to compute a lower bound of the log-likelihood ($\elbo$). We consider a mean field approximation for the latent points $q(\boldsymbol{X})$ and a joint Gaussian distribution for $q(\boldsymbol{U})$
\begin{align}
    q(\boldsymbol{U}) &= \prod_{d=1}^D \mathcal{N}(\boldsymbol{u}_d | \boldsymbol{\mu}_d, \boldsymbol{\Sigma}_d) \label{eq:prior_u}\\
    q(\boldsymbol{X}) &= \prod_{n=1}^N \prod_{q=1}^Q \mathcal{N}(x_{n,q} | m_{n,q}, s^2_{n,q}) \label{eq:prior_x}.
\end{align}
Following \cite{titsias2010bayesian} and \cite{gal2015latent}, we obtain the $\elbo$ (represented as $\mathcal{L}$) by applying Jensen's inequality with a variational distribution of the latent variables (full derivation can be found in Sec.\ 1 of Suppl.\ Material),
\begin{align}
 \log p(\boldsymbol{Y}) 
 &\geq - \underbrace{\kl(q(\boldsymbol{X})||p(\boldsymbol{X}))}_{\kl_{\boldsymbol{X}}} - \underbrace{\kl(q(\boldsymbol{U})||p(\boldsymbol{U}))}_{\kl_{\boldsymbol{U}}} \nonumber \\ & \ \ \ \ \ + \sum_{d=1}^D \int q(\boldsymbol{X})q(\boldsymbol{u}_d)p(\boldsymbol{f}_d | \boldsymbol{X},\boldsymbol{u}_d) \nonumber \\ {}& \hspace{2cm} \cdot \log p(\boldsymbol{y}_{d}|\boldsymbol{f}_{d})d\boldsymbol{X}d\boldsymbol{f}_{d}d\boldsymbol{U} \nonumber \\ &= \mathcal{L} .
 \label{eq:kl_unwrap}
\end{align}
We further marginalise $\boldsymbol{u}_d$ in the posterior distribution of $\boldsymbol{f}_d$ to obtain, 
\begin{align}
    q(\boldsymbol{f}_d | \boldsymbol{X}) &= \int p(\boldsymbol{f}_d | \boldsymbol{X}, \boldsymbol{u}_d)q(\boldsymbol{u}_d)d\boldsymbol{u}_d \\
    &= \mathcal{N}(\boldsymbol{f}_d | \boldsymbol{K}_{NM}\boldsymbol{K}_{MM}^{-1}\boldsymbol{\mu}_d, \boldsymbol{K}_{d})\label{eq:post_f} 
\end{align}
\begin{align}
    \boldsymbol{K}_d &= \boldsymbol{K}_{NN} 
    +\boldsymbol{K}_{NM}\boldsymbol{K}_{MM}^{-1}(\boldsymbol{\Sigma}_d - \boldsymbol{K}_{MM})\boldsymbol{K}_{MM}^{-1}\boldsymbol{K}_{NM}^{T}, \nonumber 
\end{align}
where $\boldsymbol{\mu}_d$ and $\boldsymbol{\Sigma}_d$ are the variational parameters and $\boldsymbol{K}_{NM}$ is the cross-covariance matrix computed over $\boldsymbol{Z}$ and $\boldsymbol{X}$. Similarly, $\boldsymbol{K}_{MM}$ as well as $\boldsymbol{K}_{NN}$ are the kernel matrices computed on $\boldsymbol{Z}$ and $\boldsymbol{X}$ respectively. Using Eq.~\eqref{eq:post_f}, we can write Eq.~\eqref{eq:kl_unwrap} as:
\begin{align}
    \mathcal{L} = &- {\kl}_{\boldsymbol{X}} - {\kl}_{\boldsymbol{U}}\label{eq:post_fin}\\&+\sum_{d=1}^D \int q(\boldsymbol{X})q(\boldsymbol{f}_d | \boldsymbol{X})\log p(\boldsymbol{y}_{d}|\boldsymbol{f}_{d})d\boldsymbol{X}d\boldsymbol{f}_{d}.\nonumber
\end{align}
To solve the integral over $\boldsymbol{X}$, we make use of Monte Carlo integration by drawing samples, $\boldsymbol{X}_i$ from $q(\boldsymbol{X})$. Hence from Eq.~\eqref{eq:post_fin}, we can write the lower bound $\mathcal{L}$ as:
\begin{align}
    \label{eq:lower_bound}
    \mathcal{L} \approx &- {\kl}_{\boldsymbol{X}} - {\kl}_{\boldsymbol{U}}\nonumber\\& + \frac{1}{N_x}\sum_{i=1}^{N_x}\sum_{d=1}^{D} \mathbb{E}_{q(\boldsymbol{f}_d|\boldsymbol{X}_i)}[\log p(\boldsymbol{y}_{d}|\boldsymbol{f}_{d})],
\end{align}
where $N_x$ corresponds to the number of samples drawn.

\subsection{Numerical quadrature}
The variational expectation over the log-likelihood, $\log p(\boldsymbol{y}_{d}|\boldsymbol{f}_{d})$ in Eq.~\eqref{eq:lower_bound} is intractable. We solve this by making use of the Gauss-Hermite quadrature \citep{liu1994note}. Hence, we follow a sampling-based approach \citep{ kingma2013auto, titsias2014doubly, rezende2014stochastic, hensman2015scalable} to compute the lower bound $\mathcal{L}$ as well as its derivatives with Gauss-Hermite quadrature. Concretely, we transform the random variables to be sampled using the re-parameterisation trick introduced in \cite{kingma2013auto}. The transformation for (vectorised) $\boldsymbol{X}$ is as follows:
\begin{align*}
\boldsymbol{X}_i = \boldsymbol{m} + \boldsymbol{s} \boldsymbol{\epsilon}_i^{(x)}, \quad \boldsymbol{\epsilon}_n^{(x)} \sim \mathcal{N}(0,I_{N \cdot Q}),
\end{align*}
where $\boldsymbol{m}$ and $\boldsymbol{s}$ contain the variational parameters from Eq.~\eqref{eq:prior_x}. For the quadrature approximation of the expectation $\mathbb{E}_{q(\boldsymbol{f}_d|\boldsymbol{X}_i)}[\log p(\boldsymbol{y}_{d}|\boldsymbol{f}_{d})]$, $\boldsymbol{f}_{d}$ (where $\boldsymbol{f}_{d} \in \mathbb{R}^{N \cdot P_d \times 1}$) can be transformed as
\begin{align*}
\boldsymbol{f}_{d}^{(j)} = \boldsymbol{a}_{d} + \boldsymbol{b}_{d} t_j,
\end{align*}
where $t_j$ is the $j$\textsuperscript{th} zero of the $J$\textsuperscript{th} order Hermite polynomial as specified by the Gauss-Hermite quadrature and $\boldsymbol{a}_d$ as well as $\boldsymbol{b}_{d}$ are specified from Eq.~\eqref{eq:post_f}:
\begin{align*}
    \boldsymbol{a}_{d} &= \boldsymbol{K}_{NM}\boldsymbol{K}_{MM}^{-1}\boldsymbol{\mu}_d \ \ \ \text{and} \ \ \ 
    \boldsymbol{b}_{d} = \sqrt{\mathrm{diag}(\boldsymbol{K}_d)} \\
    \boldsymbol{K}_d &= \boldsymbol{K}_{NN}
    +\boldsymbol{K}_{NM}\boldsymbol{K}_{MM}^{-1}(\boldsymbol{\Sigma}_d - \boldsymbol{K}_{MM})\boldsymbol{K}_{MM}^{-1}\boldsymbol{K}_{NM}^{T},
\end{align*}
where $\mathrm{diag}(\cdot)$ forms a diagonal matrix by setting all non-diagonal elements to zero and $\sqrt{\cdot}$ is element-wise (for diagonal elements).

Hence, we can approximate the expectation for the log-likelihood as a sum of $N \cdot P_d$ one-dimensional numerical quadratures,
\begin{multline}
    \mathbb{E}_{q(\boldsymbol{f}_d|\boldsymbol{X})}[\log p(\boldsymbol{y}_{d}|\boldsymbol{f}_{d})] \approx\\ \sum_{n=1}^{N}\sum_{j=1}^{J}\sum_{p=1}^{P_d} w_{j}q(f_{n,d,p}^{(j)}|\boldsymbol{X})\log p(\boldsymbol{y}_{n,d}|f_{n, d,p}^{(j)}),   \label{eq:quad_eq2}
\end{multline}
where $f_{n,d,p}^{(j)}$ indexes the $N \cdot P_d$ elements of $\boldsymbol{f}_d^{(j)}$. In our case, we take $J=3$ making Eq.~\eqref{eq:quad_eq2} a 3-point Gauss-Hermite quadrature and $w_j$ are the suitably corresponding weights. 
\subsection{Variational recognition models}
In the standard GPLVM model, there is no constraint that prevents two points which are close in data space to be embedded far apart in latent space \citep{lawrence2006local}. Moreover, the use of minibatch-based stochastic variational inference can be impractical for modest size datasets as achieving convergence can take a long time. This is because only the local parameters for a minibatch in each iteration are updated and the optimal $q(\boldsymbol{X})$ found for the other data points is ignored \citep{bui2015stochastic}. We borrow ideas from \citep{bui2015stochastic, lawrence2006local, rezende2014stochastic} and parameterise the mean and covariance of the variational distribution over $q(\boldsymbol{X})$ using neural network based recognition models. Concretely, the mean and covariance of $q(\boldsymbol{x}_n)$ are obtained as the output of two feed-forward, multi-layer perceptrons (see  Table 1 in Suppl.\ Material for more details) whose weights are trained by stochastic optimisation
\begin{align*}
q(\boldsymbol{x}_n|\boldsymbol{y}_n) = \mathcal{N}(\boldsymbol{x}_n | \mathcal{M}_{\omega_1}(\boldsymbol{y}_n), R_{\omega_2}(\boldsymbol{y}_n)^T R_{\omega_2}(\boldsymbol{y}_n)),
\end{align*}
where $\mathcal{M}$ is the mean, $R$ is the cholesky factor of the covariance, and $\omega_1$ as well as $\omega_2$ are the network weights. Therefore, $\boldsymbol{m}_n = \mathcal{M}_{\omega_1}(\boldsymbol{y}_n)$ and $\boldsymbol{S}_{n}=R_{\omega_2}(\boldsymbol{y}_n)^T R_{\omega_2}(\boldsymbol{y}_n)$. By parameterising the distribution over the latent variables with a mapping from the observations, we are introducing a constraint that encourages observations that are close in the data space to be close in the latent representation. Moreover, the use of this formulation allows for the efficient use of minibatching, thereby efficient stochastic optimisation. Specifically, the deep neural network weights, $\omega_1$ and $\omega_2$ act as global parameters that enable parameter sharing. Also, updating these parameters with respect to a data point in a minibatch also affects the latent representation of other data points. The gradients of these back-constraint parameters are obtained using the standard back-propagation algorithm. The choice of weight initialisation for the deep neural networks can affect the training of the weights \citep{sutskever2013importance}. We make use of the Xavier weight initialisation described in \citet{glorot2010understanding} for both the networks.

\subsection{Variational lower bound and stochastic optimisation}
The lower bound ($\elbo$) that needs to be optimised is valid across the data observations and hence, can be written as
\begin{align}
\mathcal{L} = &-\sum^N_{n=1}\sum^Q_{q=1}\kl(q(x_{n,q})||p(x_{n,q})) \nonumber\\&- \sum^D_{d=1}\kl(q(\boldsymbol{u}_d)||p(\boldsymbol{u}_d))\label{eq:elbo_methods}\\&+ \frac{1}{N_x} \sum^{N_x}_{i=1} \sum_{d=1}^D \sum_{n=1}^{N} \sum_{j=1}^{J} \sum_{p=1}^{P_d} w_{j}q(f_{n,d,p}^{(j)}|\boldsymbol{X_i})\nonumber\\&\hspace{4cm} \cdot\log p(\boldsymbol{y}_{n,d}|f_{n, d,p}^{(j)}).\nonumber
\end{align}
It is possible that the resulting latent embeddings may not be centred about the origin even after the model seems sufficiently optimised.
Origin-centred latent embeddings can be achieved by leveraging the idea of introducing a hyper-parameter that balances the latent channel capacity and independence constraints with reconstruction accuracy as described in \citet{higgins2017beta} (see Sec.\ 3 of Suppl.\ Material for more information).  

We can make use of a suitable stochastic optimisation technique to learn the $\elbo$. The parameters we need to optimise include the recognition model weights ($\omega_1$ and $\omega_2$), variational parameters $\boldsymbol{Z}$, $\boldsymbol{\mu}_d$, $\boldsymbol{\Sigma}_d$ and the hyper-parameters for the GP. The optimisation is done using the Adam optimiser \citep{kingma2014adam}. Adam is an adaptive learning rate method that maintains an exponentially decaying average of past gradients as well as past squared gradients. Our method allows the computation of derivatives using automatic differentiation. We make use of Theano \citep{2016arXiv160502688short} for the inference implementation and use the code released with \citep{gal2015latent} as a template for our implementation.\footnote{Source code is available at: \url{https://github.com/SidRama/Latent-GP}}
\begin{figure*}[!ht]%
    \centering
    \includegraphics[width=0.9\linewidth]{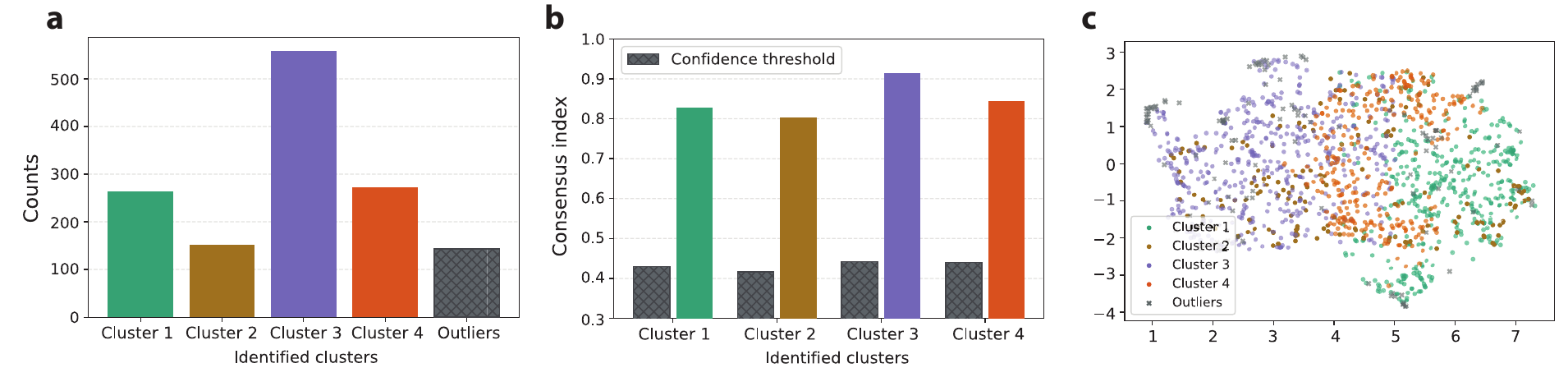} 
    
    \caption{Analysis of the resulting clusters: \textbf{(a)} number of patients in each cluster, \textbf{(b)} consensus indices, where confidence thresholds indicate Bonferroni corrected 95\% percentiles under the null model of no clustering, and \textbf{(c)} visualisation of the latent space ($Q=10$) generated by our method and projected on to a two-dimensional space using UMAP. The cluster membership is identified as described on the generated latent space.}
    \label{fig:healthcare_panel}
\end{figure*}

\section{Experiment}
\label{Experiment}
\paragraph{Clustering of clinical patient data} Personalised medicine focuses on clinical and biological characteristics of a person to optimise prediction, prevention, and treatment of diseases based on individual traits \citep{achenbach2004stratification, harvey2012future}. Diseases, such as diabetes or Parkinson's disease, manifest heterogeneous clinical symptoms that may largely vary between patients. With these diseases, for example, two or more subtypes have been identified with differing course, prognosis, and genetic associations between the subtypes  \citep{ahlqvist2018novel, kalia2015}. Thus, patient stratification based on a rich dataset of clinical and biological variables with proper statistical modelling has the potential to provide insights into the underlying disease mechanism, diagnosis, and therapy. In this experiment, we aim to identify disease subtypes by utilising heterogeneous patient records comprising of multiple likelihoods as well as noisy and missing data by embedding high-dimensional observations or feature vectors into a low-dimensional space while capturing the similarity between the observations. We show that incorporating all the covariates through the use of a composite likelihood constitutes a rigorous statistical model and yields promising results. Our experiment is summarised in Fig.~1 in the Suppl.\ Material. 

The data comprised of diagnostic disease classifications and clinical laboratory tests of patients having Parkinson's disease treated in the HUS Helsinki University Hospital, Finland. The diagnostic information comprised of International Classification of Disease codes (ICD-10) at the categorical level (first three characters) obtained during a four-year follow-up period beginning at six months prior to the first Parkinson's diagnosis. The disease codes were one-hot encoded into feature vectors and modelled with binomial likelihoods. Laboratory measurements of blood (B), erythrocytes (E), plasma (P) or fasting plasma (fP), serum (S), urea (U), and leukocytes (L) (see Figs.~6, 7 in the Suppl.\ Material) taken in the window of +/- 6 months from the first Parkinson's diagnosis were included  into feature vectors as the median over the time window. Notably, the laboratory data contained missing values. Variables expressing concentrations and percentages were modelled with Gaussian and beta likelihoods, respectively. Hence, our dataset comprised of  1400 patients with 46-dimensional feature vectors consisting of 20 binomial, 20 Gaussian and 6 beta distributed variables. Also, 10\% of the patients were held-out as test data. 

First, we assessed the optimal latent dimensionality $Q$ using the dataset. Fig.~2 in the Suppl.\ Material visualises the predictive log-likelihood on the test data for different values of $Q$. The algorithm was executed three times per dimensionality and the prediction was done using the model having the highest $\elbo$ over 1000 iterations. The highest predictive log-likelihood was obtained with the dimension $Q=10$. Hence, we selected this as the best model moving forward.

We then clustered the patient data in the latent space obtained using the selected best model. Bayesian Gaussian mixture model was used to estimate the optimal number of clusters and the cluster membership of each patient (using the scikit-learn library \citep{scikit-learn}). The maximum number of clusters was set to 20 and the result with the highest lower bound on the Gaussian mixture model evidence out of 10 initialisation runs was selected. The final number of clusters was chosen by the algorithm and we excluded clusters containing less than 5\% of patients as outliers (Fig.~\ref{fig:healthcare_panel}(a)). The resulting clustering (i.e.\ cluster label assignment) was used as a reference in the next step of the analysis and is visualised in Fig.~\ref{fig:healthcare_panel}(c) by projecting on to a two-dimensional space using UMAP \citep{mcinnes2018umap}.
\begin{figure*}[!t]%
    \centering
    \includegraphics[width=\linewidth]{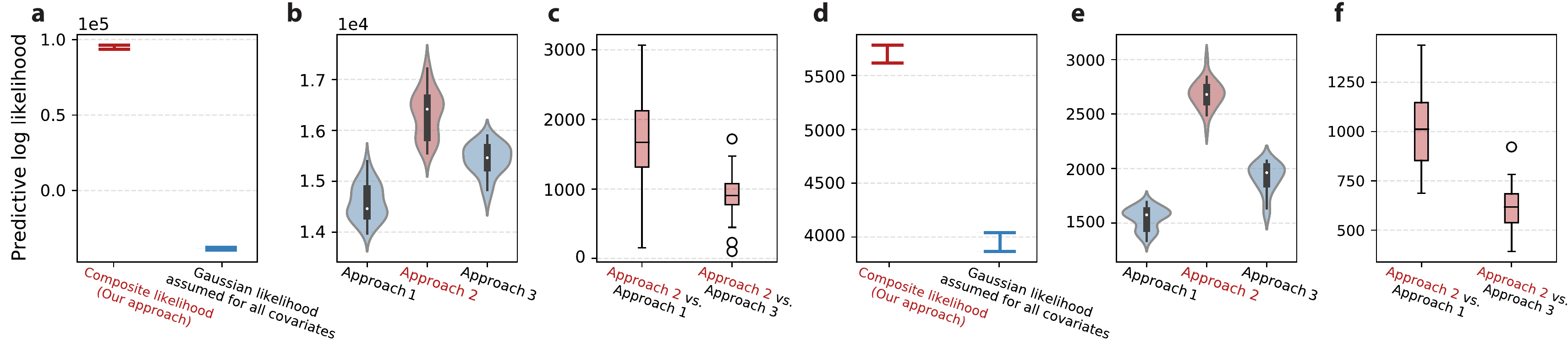}
    \caption{Predictive log-likelihoods computed on 30 sub-samples using 2-fold cross-validation (the partitions are the same across the analyses): \textbf{(a-c)} Results for clinical data from Parkinson's disease patients. \textbf{(a)} Comparison between our composite likelihood method and only Gaussian likelihood method, \textbf{(b)} violin plot comparing three approaches of estimating the Gaussian distributed covariates. \textbf{Approach 2} is our method, \textbf{(c)} a pair-wise comparison of the difference in the predictive log-likelihood between our approach and the other approaches. The paired differences are computed on the same, matched sub-samples.  \textbf{(d-f)} Visualises the same comparisons as in \textbf{(a-c)} but for the simulated data. In all panels, higher values are better.} 
    \label{fig:benhmark_panel}
\end{figure*}

\paragraph{Robustness analysis using consensus clustering} We performed a robustness analysis for the clustering by building upon the consensus clustering introduced by \citep{monti2003}. We randomly sub-sampled 50\% of the training data, ran the algorithm and applied Gaussian mixture clustering with a fixed number of clusters (using the previously obtained optimal value). Note that for selecting the optimal latent dimension, we made use of a completely separate held-out test set. To ensure that no bias was introduced, the sub-sampling in the robustness analysis was performed only on the training data (i.e., did not include the held-out test data). This routine was repeated 30 times independently. We constructed the ($N \times N$) consensus matrix, where $N$ is the original number of training samples, and where each element $i,j$ in the matrix represents the (normalised) number of times the two samples occur in the same cluster as described in \citep{monti2003}. 

The previously obtained reference clustering was used for defining the cluster membership of each entry in the consensus matrix.  This allowed the computation of the cluster-specific consensus index as the average of the entries (Fig.~\ref{fig:healthcare_panel}(b)). Moreover, we assessed the consensus indices under a null hypothesis using a permutation test. Here, the cluster membership of samples were randomly re-ordered and consensus indices were computed using the randomly re-ordered cluster memberships as the reference clustering. We defined confidence thresholds by computing the 95\textsuperscript{th} percentile over 1000 replications (with Bonferroni correction over clusters).


Also, we evaluated the differences in cluster characteristics using standard statistical tests. We computed the logarithmic odds-ratio separately for each binomial variable between the values of samples belonging to a specific cluster and the rest of the data. Similarly, for other variables, the t-statistic was applied. From Figs.~6 and 7 in the Suppl.\ Material, we can see that our method identifies clusters which appear to possess contrasting characteristics. Altogether, these results demonstrate the feasibility of our approach in finding patient subsets in a data-driven manner.

\paragraph{Benchmark and performance comparisons} We make use of 2-fold cross-validation with the predictive log-likelihood as the evaluation score to compare our method with other approaches. In other words, as described in the robustness analysis, we sub-sampled 50\% of the original data for training and then computed the predictive log-likelihoods on the remaining data. This process was repeated 30 times and the same partitions (folds) were used across all the analyses.

We compared the improvement of using multiple likelihoods (i.e.\ Gaussian, binomial or beta likelihoods depending on the covariate) against using just Gaussian likelihoods for all covariates as in the standard GPLVM. Fig.~\ref{fig:benhmark_panel}(a) compares the predictive log-likelihoods computed on the clinical dataset. As expected, the use of a composite likelihood results in a significantly higher predictive performance over all the partitions. 

Furthermore, in Fig.~\ref{fig:benhmark_panel}(b), we compare the predictive performance of three approaches specifically on the Gaussian distributed lab measurements. \textbf{Approach 1} pertains to training the model only on Gaussian distributed covariates using just the Gaussian likelihood, \textbf{Approach 2} pertains to our method of training on all covariates using their appropriate likelihoods, and \textbf{Approach 3} pertains to using a Gaussian likelihood for all the covariates (irrespective of how they may be distributed). Our method achieves higher predictive log-likelihood than standard GPLVM. 

The predictive log-likelihoods in Fig.~\ref{fig:benhmark_panel}(b) contain technical variation due to random sub-sampling. In Fig.~\ref{fig:benhmark_panel}(c), we reduce that technical variation by visualising the pair-wise differences across matched sample partitions between our method and the two other described approaches. We can see that all the differences are above zero (i.e.\ \textbf{Approach 2} has a higher predictive log-likelihood across all sub-samples). Therefore, the cross-validation analyses has shown that our method of modelling all covariates with an appropriate likelihood gives a significantly higher predictive log-likelihood (and hence captures the data generating function better) than the standard approach of assuming a Gaussian likelihood for all covariates. 

\paragraph{Demonstration on simulated data} We further demonstrated the efficacy of our method on a simulated heterogeneous dataset by performing clustering in the latent space as well as evaluating robustness and predictive accuracy. To generate the simulated dataset, we took a subset of the MNIST dataset \citep{lecun2010mnist} by choosing 3 digits and randomly sampling 400 instances of each digit class. This gives us a total of 1200 digits. The original data is in grey-scale and each pixel value ranges from 0 to 255. We re-scaled the values to the range [0, 1] and binarised the first 392 pixels of each image such that each pixel value was stochastically set to 1 in proportion to its pixel intensity \citep{salakhutdinov2008quantitative}. There are totally 784 pixels per image.
\begin{figure}[!t]%
    \centering
    \includegraphics[width=\linewidth]{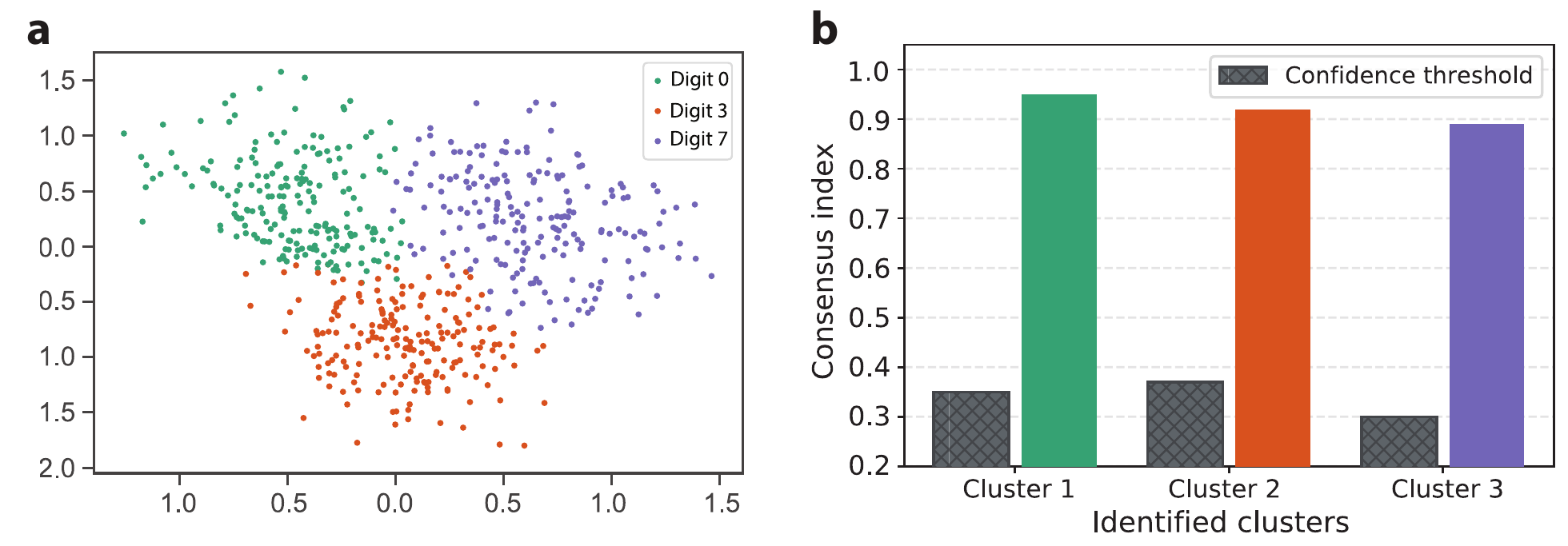} 
    \caption{Analysis of simulated data: \textbf{(a)} visualisation of the latent space ($Q=6$) generated by our method and projected onto a two-dimensional space using UMAP. The points are coloured using the known, true digit labels, and \textbf{(b)} consensus indices, where confidence thresholds indicate Bonferroni corrected 95\% percentiles under the null model of no clustering.}%
    \label{fig:MNIST_panel}
\end{figure}

We followed a similar procedure as in the clinical patient data experiment, and assigned a binomial likelihood (equivalent to Bernoulli) for the first 392 pixels (or columns) and a Gaussian likelihood for the remaining 392 pixels. The highest predictive log-likelihood was obtained with $Q=6$ which is selected as the best model for this data. Fig.~\ref{fig:MNIST_panel}(a) visualises the latent space in two dimensions coloured using the known true labels and Fig.~\ref{fig:MNIST_panel}(b) shows the results from the robustness analysis. Furthermore, we compared the numerical quadrature with the sampling based inference that utilises the marginalised distribution in Eq.~\eqref{eq:post_f}. We observed that the numerical quadrature is about two-fold more efficient in terms of computation time (Suppl.\ Fig.~11) and typically provided more robust model training (Suppl.\ Fig.~12), yet provided comparable predictive performance (Suppl.\ Fig.~11). Finally, Fig.~\ref{fig:benhmark_panel}(d-f) visualises the benchmark and performance comparisons for this dataset. Similar to the clinical dataset, our method achieves significantly better predictive likelihoods than the standard GPLVM that assumes Gaussian likelihood for all the features.
\section{Discussion and Conclusions}
\label{Conclusions}
This work proposes a generative model that is targeted to heterogeneous datasets that comprise of high-dimensional data from several disparate sources. We extend the standard GPLVM by adapting the inference framework proposed in \citep{titsias2010bayesian} and back-constraining the latent space using recognition models to produce low-dimensional embeddings of heterogeneous datasets, while preserving the similarities between the observations. We show that our method outperforms the standard GPLVM methods that are not adapted to heterogeneous likelihoods. Our approach identifies sub-groups from the heterogeneous patient data and we also demonstrate the robustness of the findings. 
This work incorporates composite likelihoods as well as sampling and numerical quadrature based variational inference to existing GPLVM techniques in the field of generative modelling and demonstrates its effectiveness.


\subsubsection*{Acknowledgements and Funding}
We would like to acknowledge the computational resources provided by Aalto Science-IT, Finland. We would also like to thank Gleb Tikhonov and Henrik Mannerstr\"om for helpful discussions and comments, Jani Salmi for data preparation, Anu Loukola for project management, and Olli Carp\'en for discussions and support. This work was supported by the Academy of Finland [292660, 313271] and Business Finland [2383/31/2015].


\bibliographystyle{abbrvnat}
\bibliography{references}

\end{document}


%
\runningtitle{Latent Gaussian process with composite likelihoods and numerical quadrature}

%
\runningauthor{Ramchandran, Koskinen, L\"ahdesm\"aki}
\onecolumn
\aistatstitle{Latent Gaussian process with composite likelihoods\\and numerical quadrature: \\ 
Supplementary Materials}

\section{Derivation of the evidence lower bound ($\elbo$)}
To obtain the $\elbo$, we can first write the log-likelihood as,
\begin{align*}
   \log p(\boldsymbol{Y}) &= \log \int p(\boldsymbol{X})p(\boldsymbol{U})p(\boldsymbol{F}|\boldsymbol{X}, \boldsymbol{U})p(\boldsymbol{Y}|\boldsymbol{F})d\boldsymbol{X}d\boldsymbol{F}d\boldsymbol{U}. 
\end{align*}
Multiplying and dividing by $q(\boldsymbol{X}, \boldsymbol{F}, \boldsymbol{U})$, we can re-write the log-likelihood as,
\begin{align}
   \log p(\boldsymbol{Y}) = \log \int \frac{q(\boldsymbol{X}, \boldsymbol{F}, \boldsymbol{U})}{q(\boldsymbol{X}, \boldsymbol{F}, \boldsymbol{U})}p(\boldsymbol{X}){}&p(\boldsymbol{U})p(\boldsymbol{F}|\boldsymbol{X}, \boldsymbol{U})p(\boldsymbol{Y}|\boldsymbol{F})d\boldsymbol{X}d\boldsymbol{F}d\boldsymbol{U}. \label{eq:log_lik}
\end{align}
The variational approximation to the posterior distribution, $q(\boldsymbol{X}, \boldsymbol{F}, \boldsymbol{U})$, can be factorised as follows:
\begin{equation}
\label{eq:posterior}
q(\boldsymbol{X}, \boldsymbol{F}, \boldsymbol{U}) = q(\boldsymbol{X})q(\boldsymbol{U})p(\boldsymbol{F}|\boldsymbol{X}, \boldsymbol{U}).
\end{equation}
Substituting into Eq.~\eqref{eq:log_lik}, we get:
\begin{align}
\log p(\boldsymbol{Y}) = \log &\int q(\boldsymbol{X})q(\boldsymbol{U})p(\boldsymbol{F}|\boldsymbol{X}, \boldsymbol{U})
\frac{p(\boldsymbol{X})p(\boldsymbol{U})p(\boldsymbol{F}|\boldsymbol{X}, \boldsymbol{U})p(\boldsymbol{Y}|\boldsymbol{F})}{q(\boldsymbol{X})q(\boldsymbol{U})p(\boldsymbol{F}|\boldsymbol{X}, \boldsymbol{U})}d\boldsymbol{X}d\boldsymbol{F}d\boldsymbol{U}.
\label{eq:log_sub}
\end{align}
Jensen's inequality relates the value of a concave (or convex) function of an integral to the integral of the concave (or convex) function \citep{jensen1906fonctions}. Assume $\varphi$ is a concave function and $X$ is a random variable. By the Jensen's inequality for a concave function, we can write:
\begin{align}
    \varphi(\E[X]) \geq \E[\varphi(X)].
    \label{eq:jensen_egs}
\end{align}
In our model, we have $\varphi = \log$. Substituting this in Eq.~\eqref{eq:jensen_egs} and for a random variable $X$, we have:
\begin{align}
    \log(\E[X]) \geq \E[\log(X)].
    \label{eq:jensen_log}
\end{align}
We can now apply the Jensen's inequality from Eq.~\eqref{eq:jensen_log} to Eq.~\eqref{eq:log_sub}: 
\begin{equation}
\label{eq:jen}
\begin{split}
    \log p(\boldsymbol{Y}) \geq &\int q(\boldsymbol{X})p(\boldsymbol{U})p(\boldsymbol{F}|\boldsymbol{X}, \boldsymbol{U}) \log \frac{p(\boldsymbol{X})p(\boldsymbol{U})p(\boldsymbol{F}|\boldsymbol{X}, \boldsymbol{U})p(\boldsymbol{Y}|\boldsymbol{F})}{q(\boldsymbol{X})p(\boldsymbol{U})p(\boldsymbol{F}|\boldsymbol{X}, \boldsymbol{U})}d\boldsymbol{X}d\boldsymbol{F}d\boldsymbol{U}.
\end{split}
\end{equation}
The Kullback-Leibler divergence \citep{kullback1951information} between $q(\boldsymbol{X})$ and $p(\boldsymbol{X})$ as well as between $q(\boldsymbol{U})$ and $p(\boldsymbol{U})$ can be written as
\begin{align*}
    \kl(q(\boldsymbol{X})||p(\boldsymbol{X})) &= \int q(\boldsymbol{X}) \log \frac{q(\boldsymbol{X})}{p(\boldsymbol{X})}d\boldsymbol{X}, \\
    \kl(q(\boldsymbol{U})||p(\boldsymbol{U})) &= \int q(\boldsymbol{U}) \log \frac{q(\boldsymbol{U})}{p(\boldsymbol{U})}d\boldsymbol{X}.
\end{align*}
Substituting the $\kl$ divergences in Eq.~\eqref{eq:jen} and unwrapping the remaining terms along the dimension $d$ (i.e.\ the dimension of the data space) from their vectorised form, we get:
\begin{align}
 \log p(\boldsymbol{Y}) \geq &-\kl(q(\boldsymbol{X})||p(\boldsymbol{X})) - \kl(q(\boldsymbol{U})||p(\boldsymbol{U})) \nonumber \\ &+ \sum_{d=1}^D \int q(\boldsymbol{X})q(\boldsymbol{u}_d)p(\boldsymbol{f}_d | \boldsymbol{X},\boldsymbol{u}_d) \cdot \log p(\boldsymbol{y}_{d}|\boldsymbol{f}_{d})d\boldsymbol{X}d\boldsymbol{f}_{d}d\boldsymbol{u}_d =\mathcal{L} .
 \label{eq:kl_unwrap}
\end{align}

\section{Likelihood model with categorical distribution}
For the categorical distribution, we make of a formulation similar to \cite{gal2015latent} which is a generalisation of the binomial distribution. In this case, the GPs produce the weights for each of the categories. We then make use of the \emph{softmax} function to get probabilities for the categories in the range of $[0,1]$. Assume all categorical variables to have the same cardinality, $K$. Hence, $P_d = K$. For the $d$\textsuperscript{th} variable of the $n$\textsuperscript{th} entry, we can write $\bar{f}_{n,d} = \{f_{n,d,1},f_{n,d,2}, ..., f_{n,d,K}\}$. Following a similar notation to \cite{gal2015latent}, we can write $y_{n,d} \sim \softmax(\bar{f}_{n,d})$, where
\begin{align*}
    \softmax(y_{n,d}=&k;\bar{f}_{n,d})=\categorical\left(\frac{\exp(f_{n,d,k})}{\sum_{k'=1}^{K}\exp(f_{n,d,k'})}\right),
\end{align*}
and $\categorical$ corresponds to the categorical distribution (or generalised Bernoulli distribution).

\section{Origin-centred latent representations}
From Eq.~$(12)$ in the main paper, we can write the lower bound $\mathcal{L}$ with our suggested modification as follows:
\begin{align}
    \mathcal{L} \approx -\eta\overbrace{\kl(q(\boldsymbol{X})||p(\boldsymbol{X}))}^{\kl_{\boldsymbol{X}}} &- \overbrace{\kl(q(\boldsymbol{U})||p(\boldsymbol{U}))}^{\kl_{\boldsymbol{U}}} + \frac{1}{N_x}\sum_{i=1}^{N_x}\sum_{d=1}^{D} \mathbb{E}_{q(\boldsymbol{f}_d|\boldsymbol{X}_i)}[\log p(\boldsymbol{y}_{d}|\boldsymbol{f}_{d})],
    \label{eq:elbo_vectorised_hyper}
\end{align}
where $\eta$ is the new hyper-parameter. If $\eta = 1$, we get the same equation as Eq.~$(12)$ in the main paper. Using $\eta > 1$ results in embeddings that are more centred about the origin but with qualitatively consistent results (i.e.\ qualitatively consistent with the hyper-parameter $\eta = 1$). \\

\noindent To obtain a centred embedding, the hyper-parameter $\eta$ must be tuned. We have also observed that as the value of $\eta$ begins to increase to larger values, the clustering structure begins to disappear as the $\kl$ term associated with $X$ (i.e.\ $\kl_{\boldsymbol{X}}$ in Eq.~\eqref{eq:elbo_vectorised_hyper}) begins to dominate and the optimisation tries to move the latent points closer to zero while making them appear to be a sample from the standard normal distribution. \\

\noindent Hence, we can infer that incorporating a weight on $\kl_{\boldsymbol{X}}$ (i.e.\ $\eta>1$) results in latent embeddings that are generally centred around the origin with results that are qualitatively consistent with Eq.~$(12)$ in the main paper (i.e.\ $\eta=1$).

\section{Ablation study}
We demonstrate that numerical quadrature has a comparable predictive performance to the standard sampling based inference, while having a significantly lower run time. We followed an approach similar to the `Benchmark and performance comparisons' described in Section 3 of the main manuscript. We performed a 2-fold cross-validation with the predictive log-likelihood as the evaluation score to compare between the two techniques. This was done by sub-sampling $50\%$ of the original data for training and then computing the predictive log-likelihoods on the remaining data. This process was repeated 30 times and the same folds were used for both analyses. In all the runs we used $Q=2$. We can see from Fig.~\ref{fig:pred_compare_time} that quadrature takes half the amount of time for each run. Fig.~\ref{fig:elbo_quad_sampling} visualises the trace plot of the ELBOs.

\section{Dimensionality reduction techniques}
Most dimensionality reduction techniques focus on preserving the distances between nearby objects than objects that are further apart (i.e.\ local distance preservation). However, it is also important to ensure that objects that are further apart in data space are also kept apart in the reduced space (i.e. dissimilarity preservation). Unfortunately, in most cases it is not possible to achieve both. In this work, we focus on feature projection which is the transformation of data from a high-dimensional space to lower dimensional manifold. There are many popular algorithms for dimensionality reduction such as PCA \citep{jolliffe2011principal}, kernel PCA \citep{Scholkopf1997}], locally linear embedding \citep{roweis2000nonlinear}, Isomap \citep{tenenbaum2000global}, GPLVM \citep{lawrence2004gaussian}, etc. These algorithms are suitable in the homogeneous data setting and do not necessarily extend well to a heterogeneous data setting. We build upon the GPLVM to support a heterogeneous data setting.

PCA tries to identify an orthogonal linear projection of the data along the direction of maximum variance. Classical PCA has some shortcomings. It is not probabilistic as it has no likelihood model for the observed data. Moreover, computation of the covariance matrix and its associated eigendecomposition can be computationally intensive for large datasets with high dimensionality \citep{prasad2008}. Another issue with classical PCA is that it cannot handle missing data properly and is not robust to outliers \citep{kambhatlaL1997}. Hence, it is not suitable in a generative model setting.

Probabilistic principal component analysis (PPCA), proposed by \citep{tipping1999probabilistic}, is a generalisation of the classical PCA that tries to overcome its shortcomings. It incorporates a probabilistic model and obtains a linear projection by maximising the likelihood. Assume a $D$-dimensional dataset $\boldsymbol Y$ of $N$ points, i.e.,  $\boldsymbol Y = \{\boldsymbol y_n\}_{n=1}^N$ such that $\boldsymbol y_n \in \mathbb{R}^D $ and is centred. We can denote each latent variable corresponding to each data point as $\boldsymbol x_n \in \mathbb{R}^Q$ such that $Q \leq D$. Also, let $ \boldsymbol W \in \mathbb{R}^{D\times Q}$ denote the principal axes or weights. We can write the likelihood for an individual data point as
\begin{align*}
    p(\boldsymbol y_n | \boldsymbol W, \sigma^2) &= \int p(\boldsymbol y_n | \boldsymbol x_n , \boldsymbol W, \sigma^2)p(\boldsymbol x_n)d\boldsymbol x_n \\
    p(\boldsymbol x_n) &= N(\boldsymbol x_n | 0, \boldsymbol I) \\
    p(\boldsymbol y_n | \boldsymbol x_n, \boldsymbol W, \sigma^2) &= N(\boldsymbol y_n | \boldsymbol W \boldsymbol x_n, \sigma^2\boldsymbol{I}_D)
\end{align*}
where $\sigma^2$ is the noise variance and we assume an isotropic Gaussian noise model. To solve for $\boldsymbol W$ we assume that $\boldsymbol y_n$ is i.i.d.\ and maximise the likelihood for all data points
\begin{align*}
    p(\boldsymbol Y | \boldsymbol W, \sigma^2) = \prod_{n=1}^{N}p(\boldsymbol y_n | \boldsymbol W\boldsymbol x_n, \sigma^2\boldsymbol{I}_D).
\end{align*}
Marginalising out the latent variables, the distribution for each point can be written as
\begin{align*}
    \boldsymbol y_n \sim N(0, \boldsymbol{WW}^T + \sigma^2\boldsymbol I_D).
\end{align*}
The parameters are optimised to obtain the maximum likelihood. We can say that the classical PCA is a limiting case of PPCA when the covariance becomes infinitesimally small, i.e., $\sigma^2 \rightarrow 0$.

\citet{lawrence2004gaussian} proposed the Gaussian process latent variable model (GPLVM) as a generalisation of PPCA. It is an unsupervised learning algorithm that extends PPCA by making use of a less restrictive covariance function that allows for non-linear mappings. Building upon the derivation of PPCA and following \citet{lawrence2004gaussian}, we can obtain the GPLVM formulation. 

A prior distribution is defined for $\boldsymbol W$ as $p(\boldsymbol W) = \prod_{i=1}^D N(\boldsymbol w_i | 0, \alpha^{-1} \boldsymbol I)$. From the PPCA derivation, instead of integrating out the latent variables $\boldsymbol X$, the principal axes $\boldsymbol W$ is marginalised to give the marginal likelihood for $\boldsymbol Y$,
\begin{align*}
    p(\boldsymbol Y | \boldsymbol X, \sigma) &= \frac{1}{(2\pi)^{\frac{DN}{2}}|\boldsymbol K|^{\frac{D}{2}}}\exp{(-\frac{1}{2}\Tr(\boldsymbol K^{-1}\boldsymbol{YY}^T))},
\end{align*}
where $\Tr$ corresponds to the trace of a matrix and $\boldsymbol K = \alpha \boldsymbol{XX}^T + \sigma^2 \boldsymbol I$. Hence, the marginal likelihood that is being optimised can be interpreted as the product of $D$ independent Gaussian processes where $\boldsymbol K$ is given by the linear covariance function. Therefore to obtain the GPLVM formulation, a non-linear covariance function is introduced which corresponds to a non-linear mapping from latent space to data space. A common choice for the process prior is the radial basis function (RBF) kernel. The marginal likelihood is jointly marginalised with respect to the latent variables, $\boldsymbol X$ and other parameters. In this study, we make use of the automatic relevance determination radial basis function (ARD RBF) kernel that allows for separate length scales in each dimension of the latent space. \\

Hence, the optimisation problem can be written as,
\begin{align*}
    \{{\hat{\boldsymbol{X}}, \hat{\boldsymbol{\theta}}}\} = \argmax_{\boldsymbol{X},\boldsymbol{\theta}}p(\boldsymbol{Y}|\boldsymbol{X},
    \boldsymbol{\theta})
\end{align*}
where $\boldsymbol{\theta}$ corresponds to the kernel parameters and $\hat{\boldsymbol{X}}$ as well as $\hat{\boldsymbol{\theta}}$ corresponds to the optimal values of the latent variables and kernel parameters, respectively.

\section{Runtime performance}
The mean wall clock times (on an Intel Xeon E5 processor) for each repetition of 1000 epochs in Fig.~4 of the main manuscript using the simulated data was as follows:
\begin{itemize}
    \item Approach 1: 1.3 hours
    \item Approach 2 (our method): 1.35 hours
    \item Approach 3: 1.5 hours
\end{itemize}

Also, Suppl.\ Fig.~11 shows a comparison of the wall-clock time for 1000 epochs between numerical quadrature and sampling based inference in our model with the simulated data. Our results indicate that the numerical integration makes the inference about two-fold faster, and overall our model is approximately as fast as the Gaussian GPLVM. We had sub-sampled the MNIST to maintain a short computation time.

\section{Neural network architecture}
The mean and covariance of the variational distribution over $q(\boldsymbol{X})$ are parameterised by neural networks. In our experiment with clinical patient data, we utilised simple feedforward multilayered perceptrons (MLPs) as the recognition models. Concretely, we made use of two separate MLPs for the mean and covariance respectively. The hyperparameters for the networks are reported in Table~\ref{table:nnet_spec}.
\begin{table}[!h]
\begin{center}
\begin{adjustbox}{width=0.8\textwidth}
\begin{tabular}{ |c||c|c|c| } 
\hline
 & Hyperparameter & Value \\
\hline
\multirow{7}{6em}{Mean} & Dimensionality of input & 46 \\ 
& Number of hidden layers & 1 \\ 
& Width of hidden layer & 30 \\ 
& Activation function of the hidden layer & TanH \\
& Dimensionality of output & $\mathcal{Q}$ \\
& Activation function of output layer & Linear \\
& Weight initialisation & Xavier initialisation \citep{glorot2010understanding} \\
\hline
\multirow{7}{6em}{Covariance} & Dimensionality of input & 46 \\
& Number of hidden layers & 1 \\
& Width of hidden layer & 30 \\
& Activation function of the hidden layer & TanH \\
& Dimensionality of output & $\mathcal{Q}$ \\
& Activation function of output layer & Sigmoid \\
& Weight initialisation & Xavier initialisation \citep{glorot2010understanding}\\
\hline
\end{tabular}
\end{adjustbox}
\end{center}
\caption{Hyperparameters used in the recognition models for the clinical patient dataset.}
\label{table:nnet_spec}
\end{table}
\newpage

\section{Supplementary figures}
\begin{figure}[ht]
\begin{center}
\includegraphics[width=0.7\linewidth]{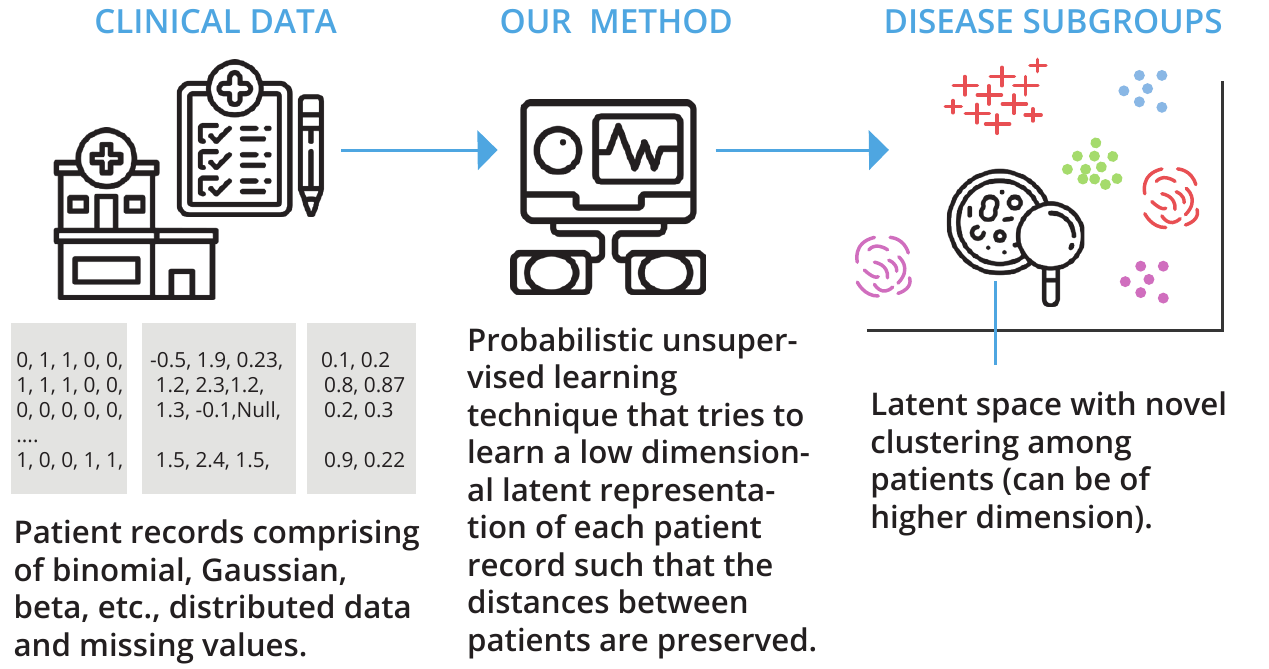}
\caption{An overview of our unsupervised generative model for disease stratification.}
\label{fig:detail_overview}
\end{center}
\end{figure}

\begin{figure}[ht]%
    \centering
    \includegraphics[width=.6\linewidth]{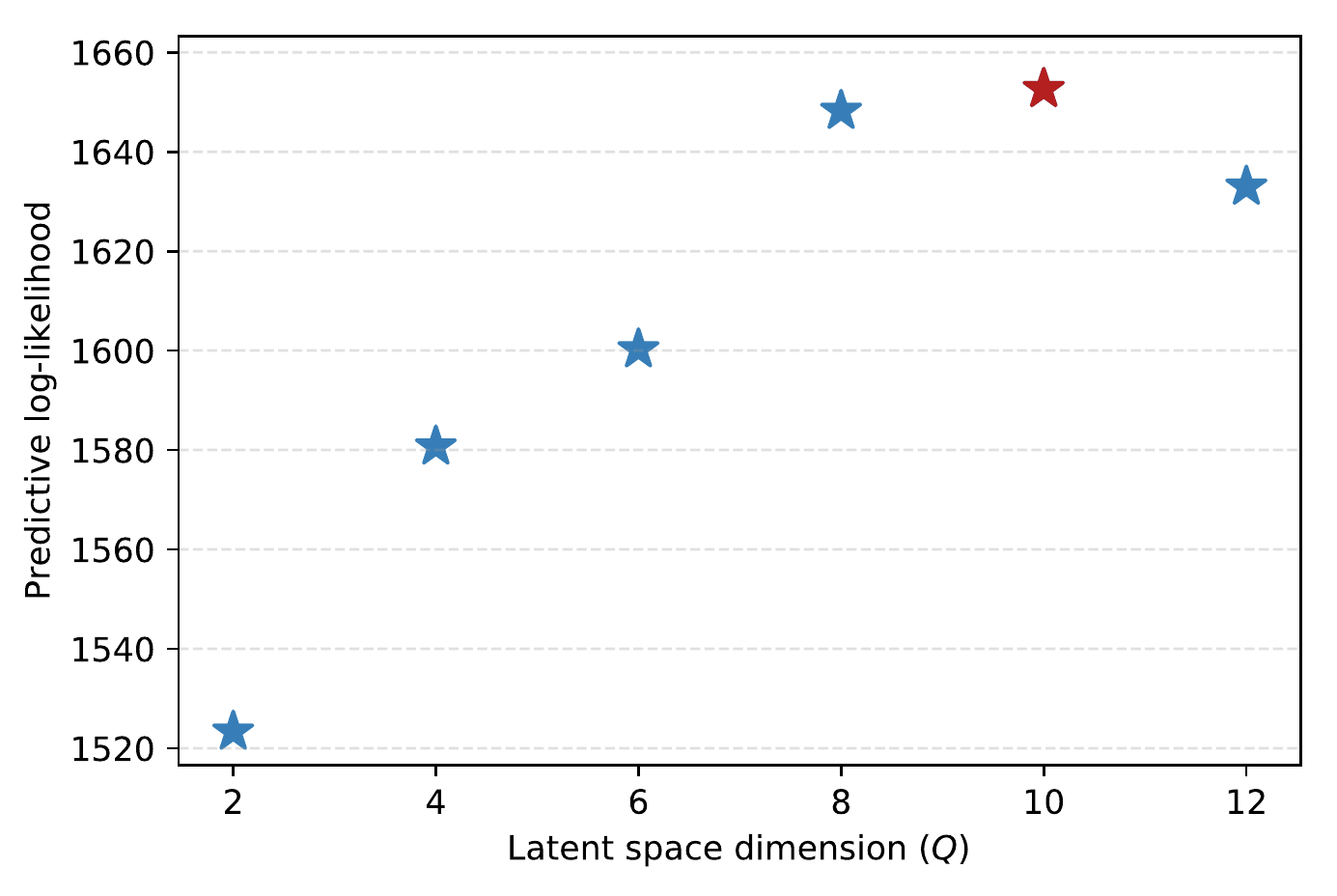}
    \caption{Assessment of optimal latent dimensionality using the predictive log-likelihood on the held-out test data for different latent space dimensions, $Q$ in the clinical patient dataset experiment. The best predictive log-likelihood among three runs is shown.}
    \label{fig:pred_lik_box}
\end{figure}
\begin{figure}[ht]
  \begin{center}
    \includegraphics[width=0.6\textwidth]{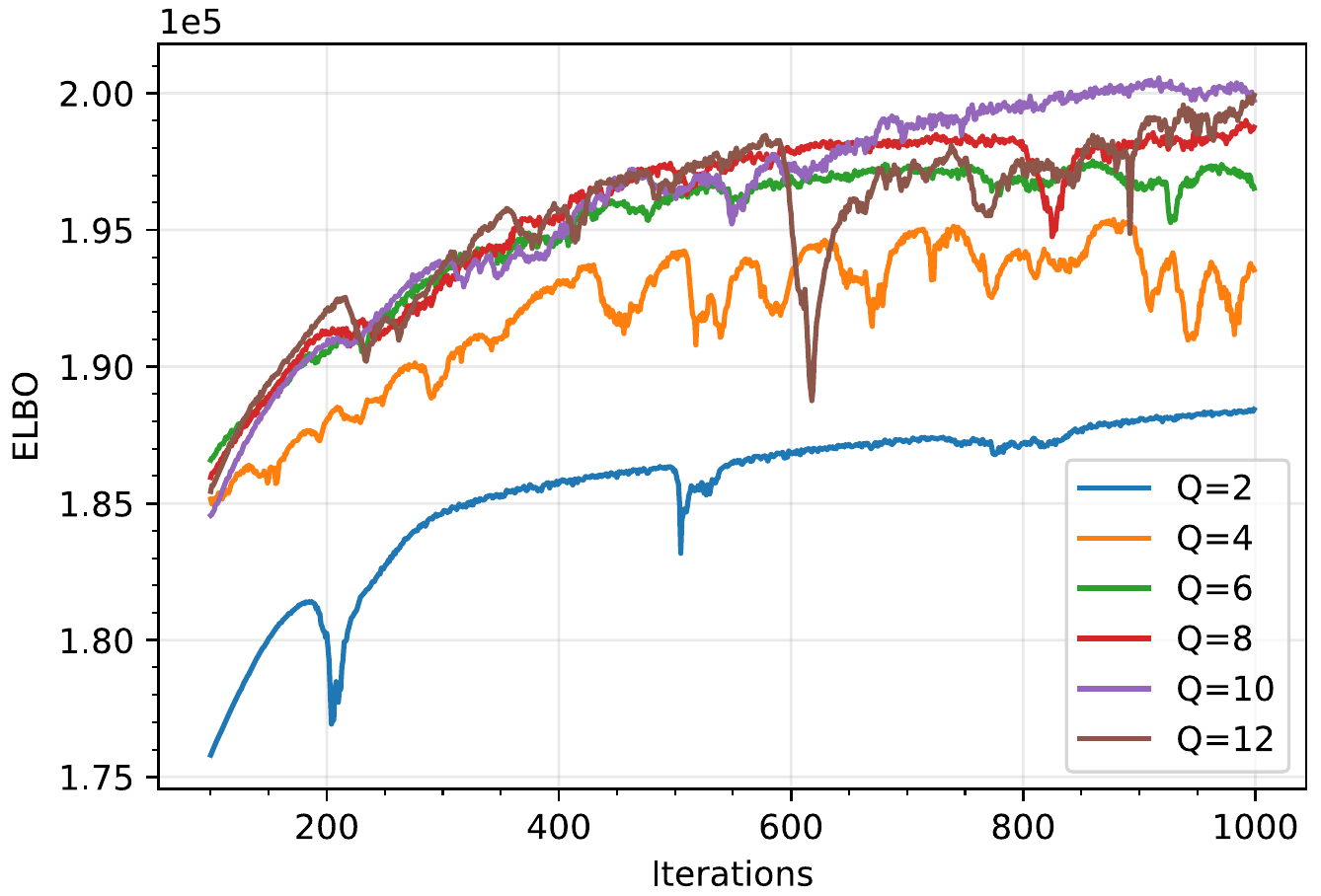}
    \caption{Trajectories of the evidence lower bounds ($\elbo$) from the best optimisation run for each latent dimension in Fig.\ \ref{fig:pred_lik_box}.}
    \label{fig:elbos}
  \end{center}
\end{figure}

\begin{figure}[!h]
  \begin{center}
    \includegraphics[width=0.6\textwidth]{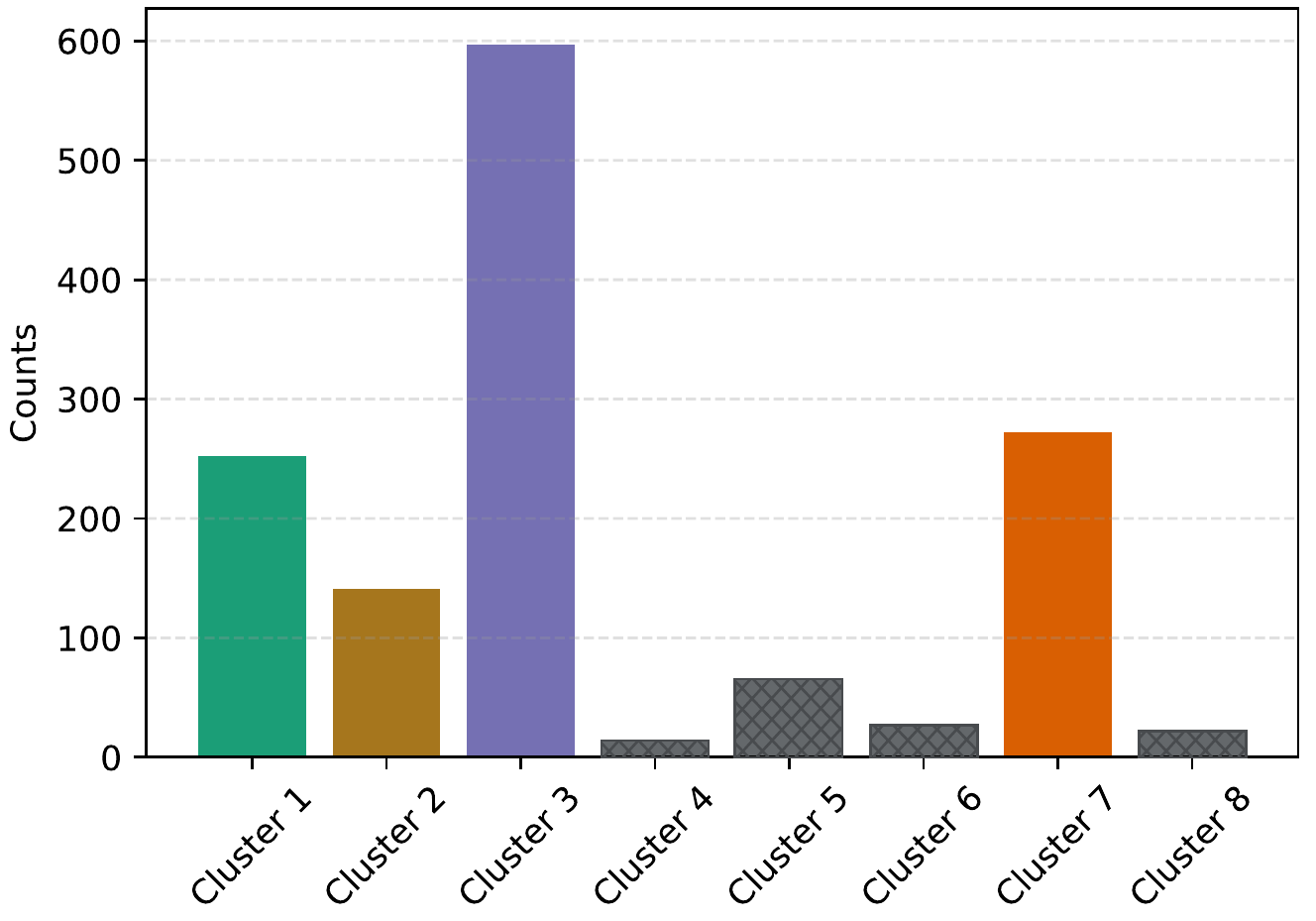}
    \caption{Visualisation of the number of patients assigned to each cluster. The optimal number of clusters and cluster membership was obtained using the described method. This figure is similar to Fig.~3(a) in the main manuscript, but includes the clusters deemed as outliers.}
    \label{fig:clusters}
  \end{center}
\end{figure}

\begin{figure}[ht]
  \begin{center}
    \includegraphics[width=0.6\textwidth]{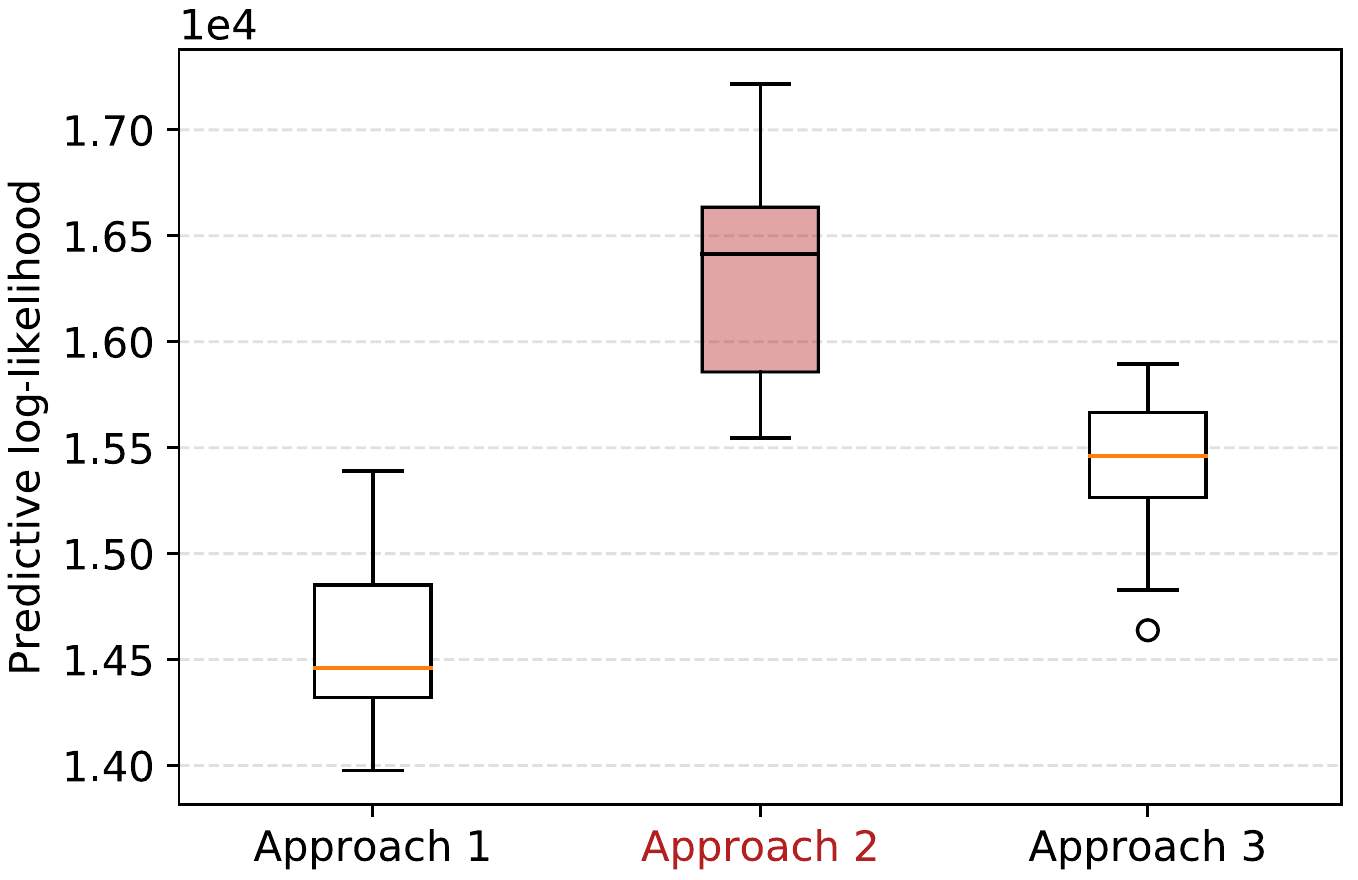}
    \caption{A box plot comparing three approaches of estimating the Gaussian distributed data similar to Fig.~4(b) in the main manuscript. \textbf{Approach 2} is our method. The predictive log-likelihood is computed on 30 sub-samples using 2-fold cross-validation (the partitions are the same across the analyses).}
    \label{fig:box_approaches}
  \end{center}
\end{figure}

\begin{figure}[!h]
\centering
\includegraphics[width=0.9\textwidth]{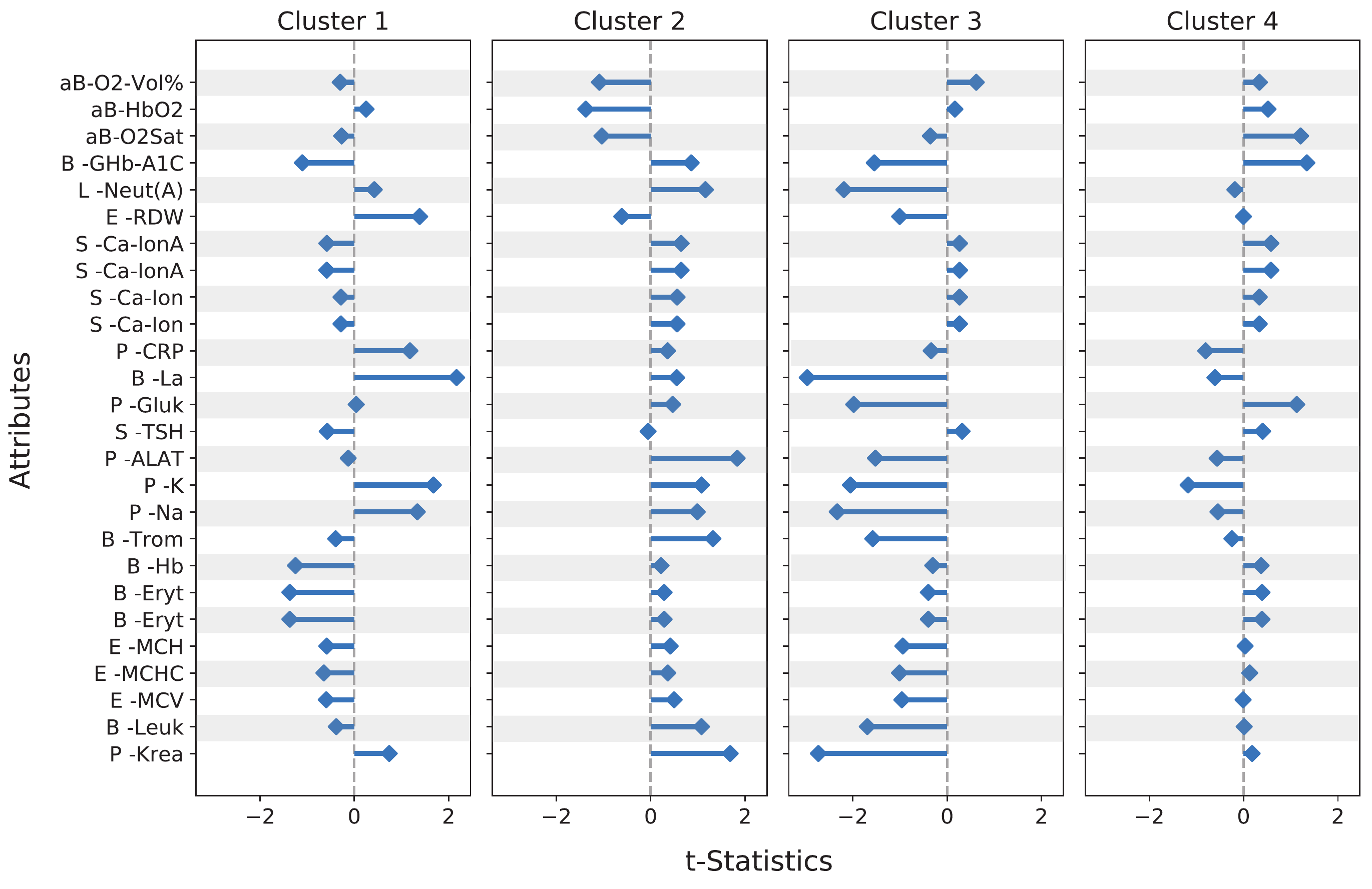}
\caption{Evaluation of cluster characteristics using t-statistics in the clinical patient dataset.}
\label{fig:tstats}
\end{figure}

\begin{figure}[!h]
\centering
\includegraphics[width=0.9\textwidth]{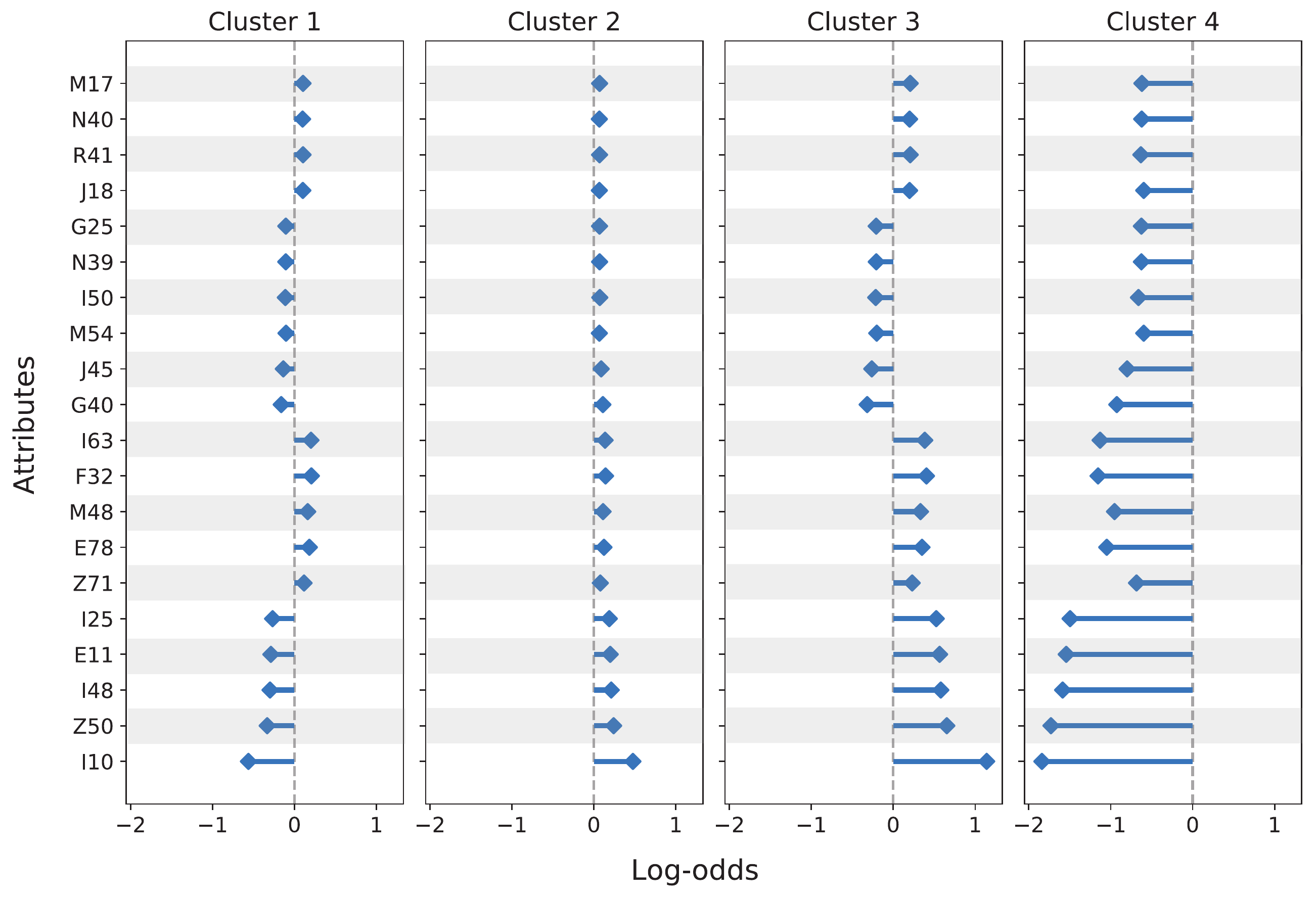}
\caption{Evaluation of binomial cluster characteristics using log-odds ratio in the clinical patient dataset.}
\label{fig:odds_ratios}
\end{figure}

\begin{figure}[!h]
\centering
\includegraphics[width=0.6\textwidth]{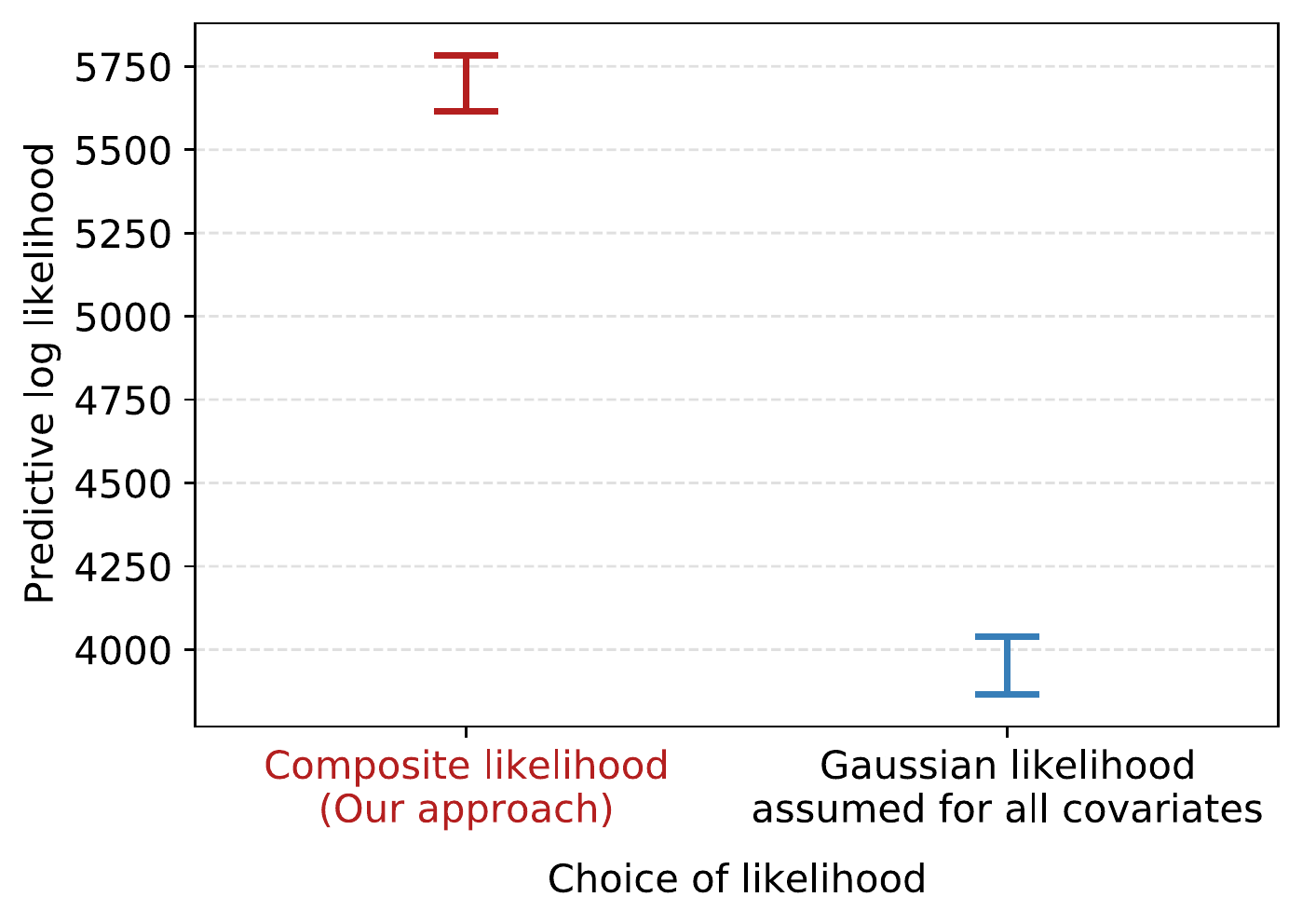}
\caption{Comparison between our composite likelihood method and only Gaussian likelihood method for the simulated dataset.}
\label{fig:MNIST_lik_comparison}
\end{figure}

\begin{figure}[!h]
\centering
\includegraphics[width=0.6\textwidth]{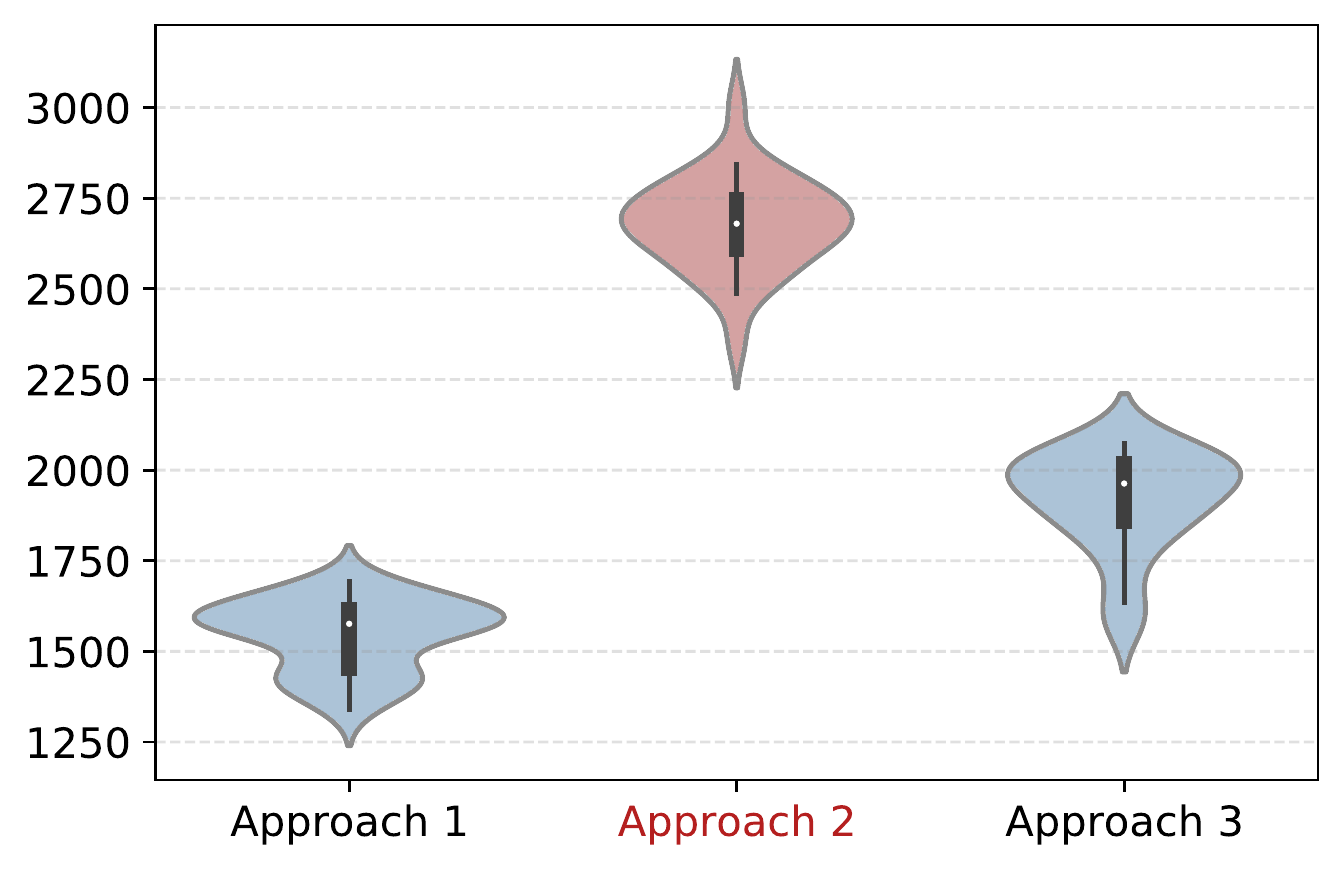}
\caption{Violin plot comparing three approaches of estimating the Gaussian distributed covariates for the simulated dataset. \textbf{Approach 2} is our method.}
\label{fig:MNIST_violin_plot}
\end{figure}

\begin{figure}[!h]
\centering
\includegraphics[width=0.6\textwidth]{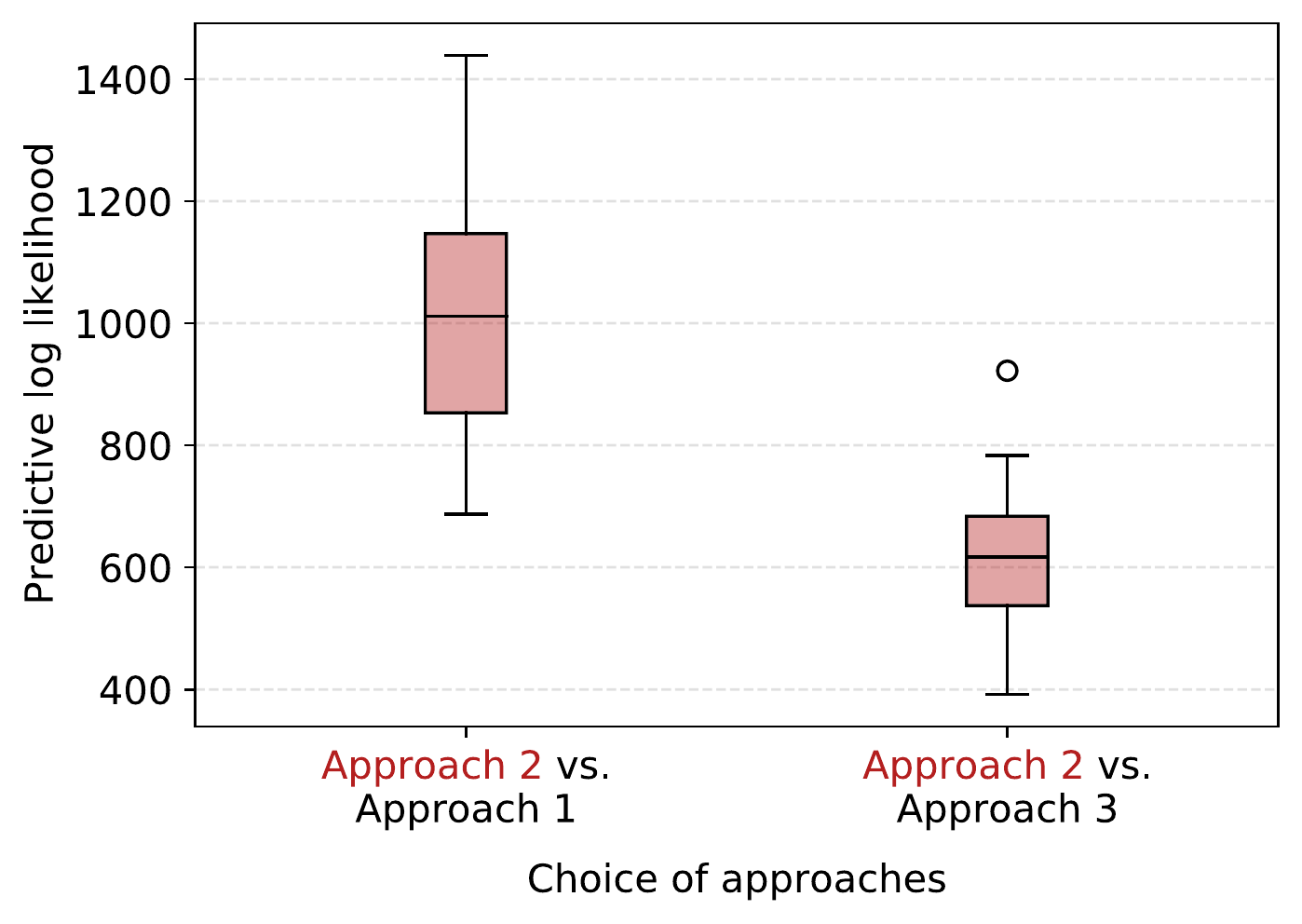}
\caption{A pair-wise comparison of the difference in the predictive log-likelihood between our approach and the other approaches for the simulated dataset. The paired differences are computed on the same, matched sub-samples.}
\label{fig:MNIST_comparison}
\end{figure}

\begin{figure}[ht]%
    \centering
    \includegraphics[width=.6\linewidth]{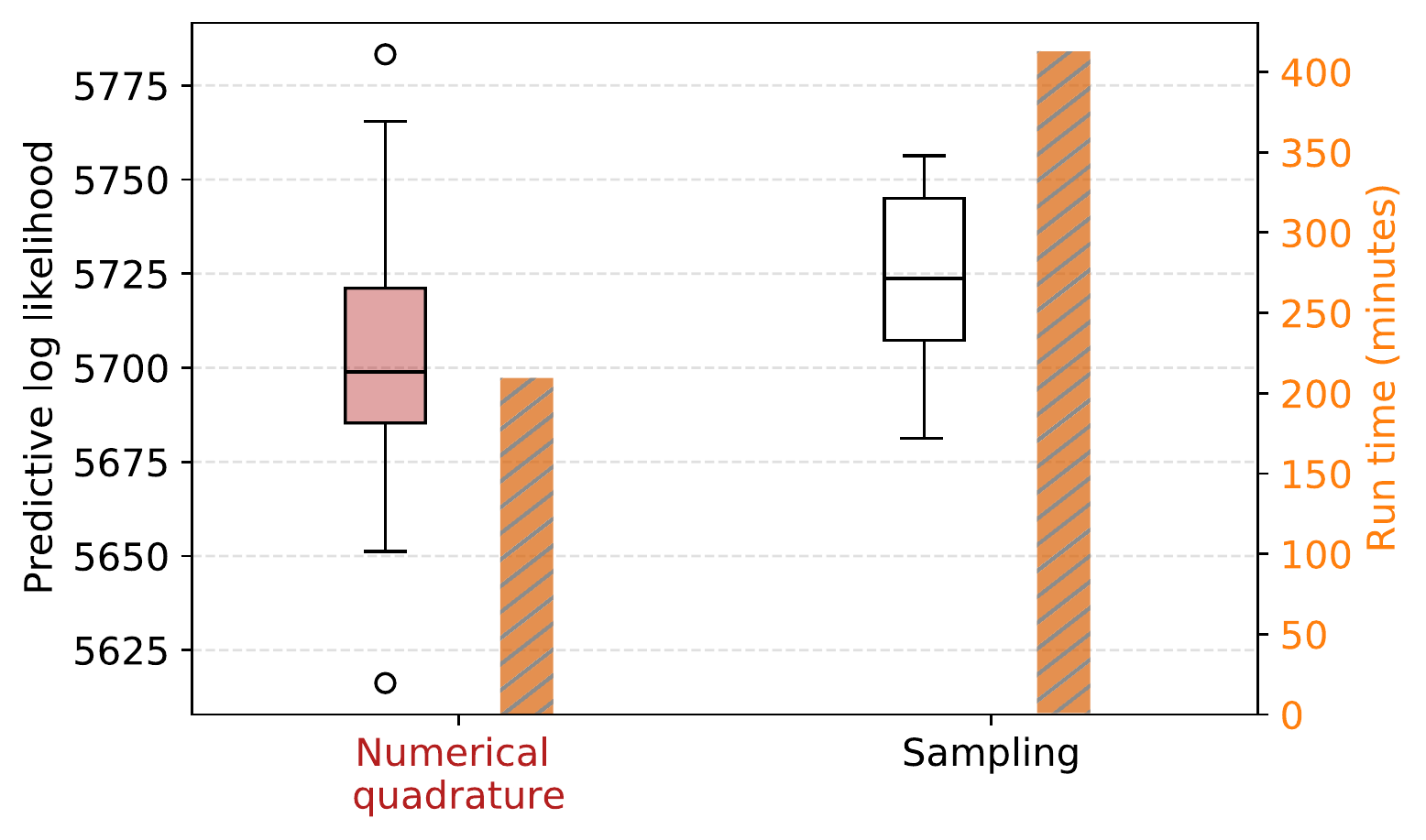}
    \caption{Comparison of the log predictive likelihood achieved by using either quadrature or sampling based inference for the simulated dataset. The wall-clock run times for 1000 epochs are shown on right.}
    \label{fig:pred_compare_time}
\end{figure}
\begin{figure}[ht]%
    \centering
    \includegraphics[width=.6\linewidth]{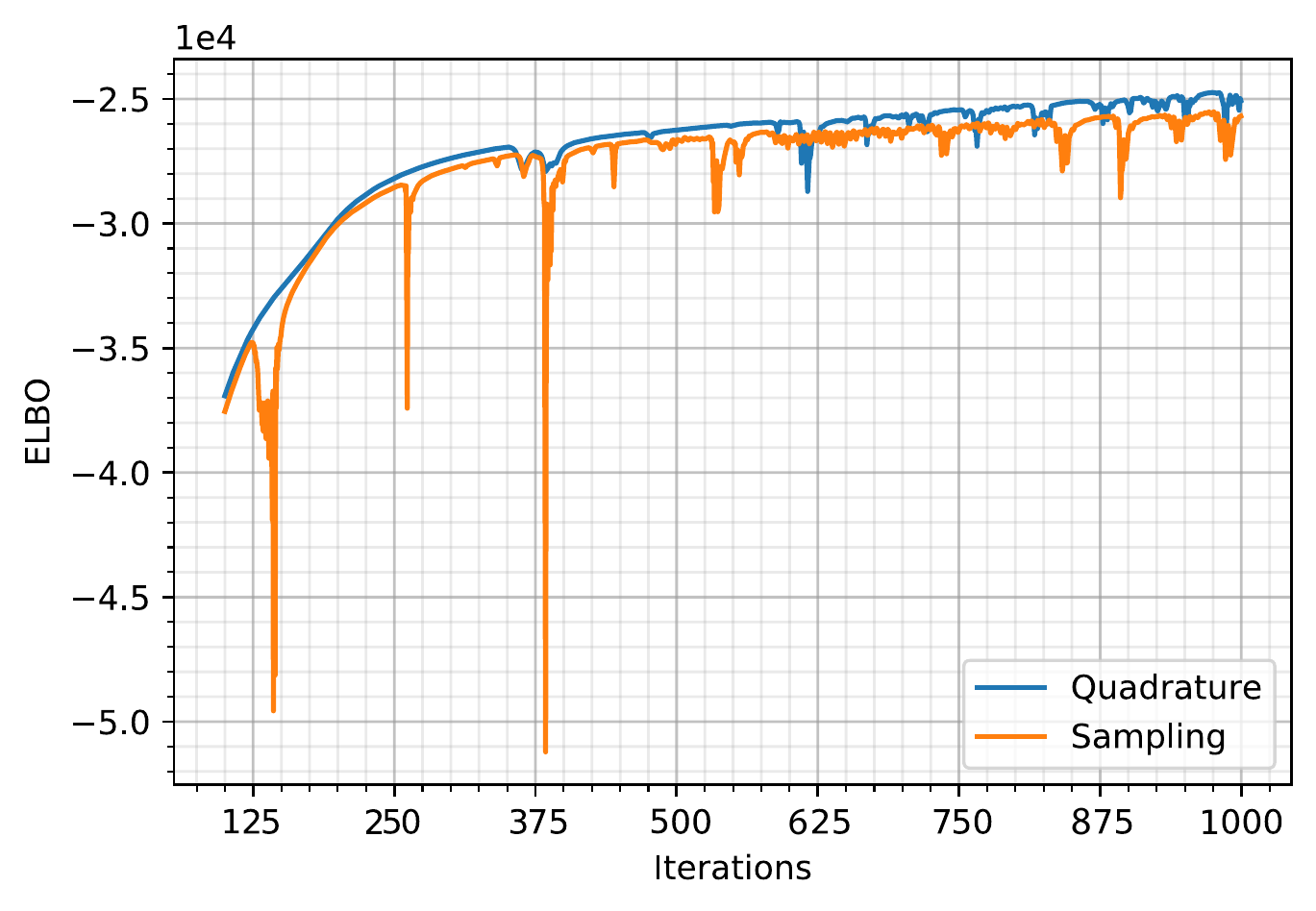}
    \caption{Trajectories of the evidence lower bounds ($\elbo$) for quadrature and sampling based inference with $Q=2$ for the simulated dataset.}
    \label{fig:elbo_quad_sampling}
\end{figure}

\begin{figure}[t]%
    \centering
    \includegraphics[width=.4\linewidth]{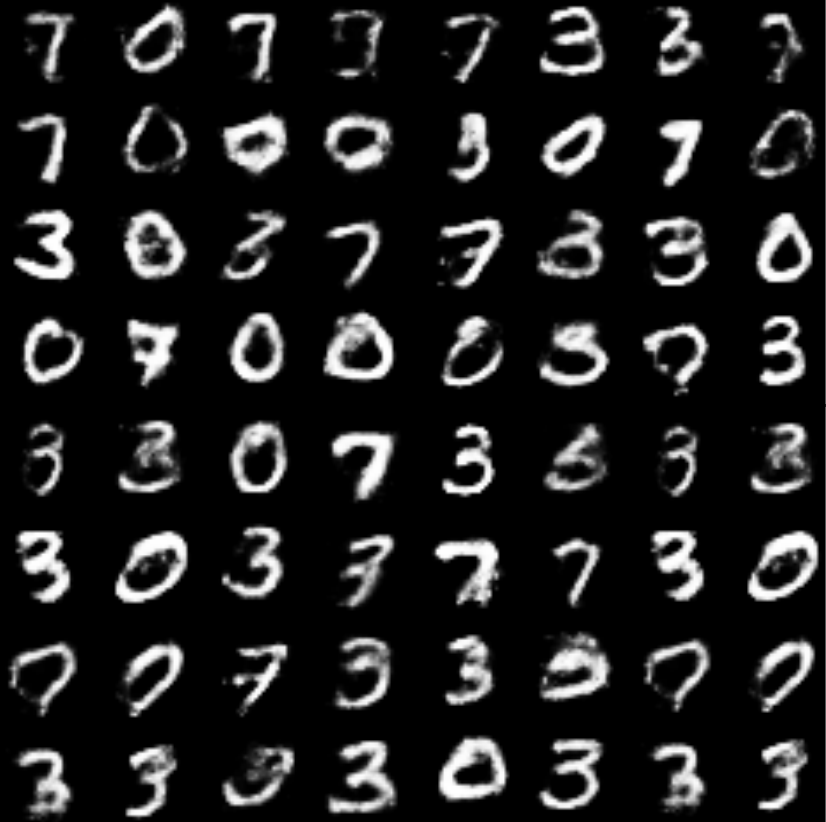}
    \caption{Reconstructions obtained in the simulated dataset experiment.}
    \label{fig:mnist_reconstruction}
\end{figure}
\FloatBarrier
\bibliographystyle{abbrvnat}
\bibliography{references}